\documentclass[3p,12pt]{elsarticle}

\usepackage{amssymb}
\usepackage{amsthm}
\usepackage{amsfonts,amstext,amsmath,amssymb}
\usepackage{hyperref}
\usepackage{multirow}
\usepackage{color}
\usepackage{subfig}
\usepackage{url}

\newcommand{\new}[1]{{\color{black}{#1}}}

\newcommand{\bx}{\boldsymbol{x}}
\newcommand{\bp}{\boldsymbol{\pi}}
\newcommand{\btheta}{\boldsymbol{\theta}}

\def\somega#1{\{\omega_#1\}}
\newcommand{\acc}{\textsf{acc}}
\newcommand{\conf}{\textsf{conf}}

\journal{International Journal of Approximate Reasoning}

\begin{document}

\begin{frontmatter}
\title{Lymphoma segmentation from 3D PET-CT images using a deep evidential network}

\author[utc,litis]{Ling Huang} \author [litis]{Su Ruan}
\author [chb]{Pierre Decazes}
\author [utc,iuf]{Thierry Den{\oe}ux}

\address[utc]{Heudiasyc, CNRS, Universit\'e de technologie de Compi\`egne, Compi\`egne, France}

\address[litis]{Quantif, LITIS, University of Rouen Normandy, Rouen, France}

\address[chb]{Department of Nuclear Medicine, Henri Becquerel Cancer Center, Rouen, France}

\address[iuf]{Institut universitaire de France, Paris, France}

\begin{abstract}
An automatic evidential segmentation method based on Dempster-Shafer theory and deep learning is proposed to segment lymphomas from three-dimensional Positron Emission Tomography (PET) and Computed Tomography (CT) images. The architecture is composed of a deep feature-extraction module and  an evidential  layer. The feature extraction module uses an encoder-decoder framework to extract semantic feature vectors from 3D inputs. The evidential layer then uses prototypes in the feature space to compute  a belief function at each voxel quantifying the uncertainty about the presence or absence of a lymphoma at this location. Two evidential layers are compared, based on different ways of  using distances to  prototypes for computing mass functions. 
The whole model is trained end-to-end by minimizing the Dice loss function. The proposed combination of deep feature extraction and evidential segmentation is shown to outperform the baseline UNet model as well as three other state-of-the-art models on a dataset of 173 patients.  
\end{abstract}


\end{frontmatter}



\section{Introduction}
\label{intro}

\begin{figure}
\includegraphics[width=\textwidth]{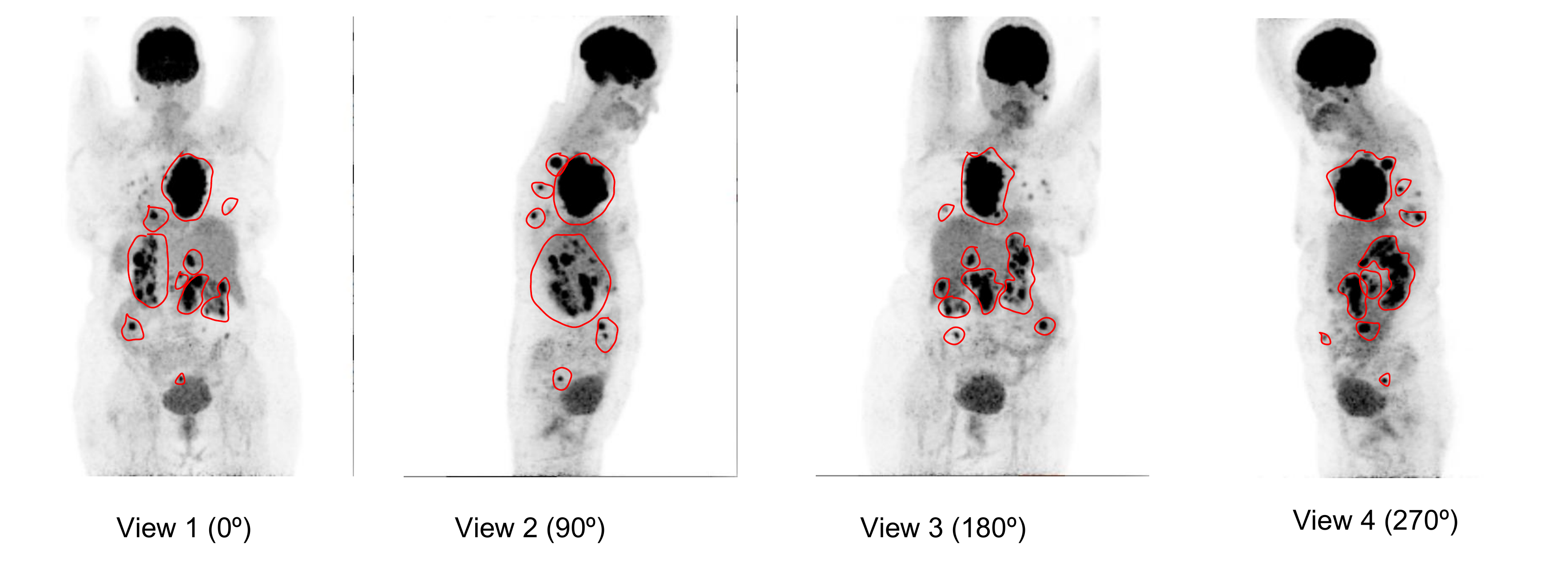}
\caption{Example of a patient with lymphomas in 3D PET-CT views. The lymphomas are the dark areas circled in red.}
\label{fig1}
\end{figure}

Positron Emission Tomography - Computed Tomography (PET-CT) scanning is an effective imaging tool for lymphoma segmentation with application to clinical diagnosis and radiotherapy planning. The standardized uptake value (SUV), defined as the measured activity normalized for body weight and injected dose to remove variability in image intensity between patients, is widely used to locate and segment lymphomas thanks to its high sensitivity and specificity to the metabolic activity of tumors \cite{jhanwar2006role}. However, PET images have a low resolution and suffer from the partial volume effect blurring the contours of objects  \cite{zaidi2010pet}. For that reason, CT images are usually used jointly with PET images because of their anatomical feature-representation capability and high resolution. Figure \ref{fig1} shows 3D PET-CT views of a lymphoma patient. The lymphomas are marked in black as well as the brain and the bladder. As we can see from this figure, lymphomas vary in intensity distribution, shape, type, and number. 

\paragraph{Approaches to lymphoma segmentation} Techniques for lymphoma segmentation can be divided into three classes: SUV-based, region-growing-based and deep learning-based methods.
For PET images, it is common to segment lymphomas with a set of fixed SUV thresholds. The so-called SUV-based methods \cite{ilyas2018defining}\cite{eude2021reproducibility} are fast but lack flexibility in boundary delineation and requires domain knowledge to locate the region of interest. Region-growing-based methods \cite{onoma2014segmentation}\cite{hu2019detection} have been proposed to optimize boundary delineation by taking texture and shape information into account. By setting the specific growing function and stopping condition, the tumor region  grows step by step until it reaches the stopping condition. However, those methods still need clinicians to locate the seeds for region growing \cite{onoma2014segmentation} and they are time-consuming, especially when applied to  3D  images. Lymphoma segmentation with deep learning has become a popular research topic thanks to its high feature representation ability \cite{li2019densex}\cite{hu2020lymphoma}. 

\paragraph{Deep-learning-based methods} Long et al. \cite{long2015fully} were the first authors to show that a  fully convolutional network (FCN) could be trained end-to-end for semantic segmentation, exceeding the state-of-the-art when the paper was published. UNet \cite{ronnebergerconvolutional}, a successful modification and extension of FCN, has become the most popular model for medical image segmentation in recent years. Driven by different tasks and datasets, several extended and optimized variants of UNet have been proposed for medical image segmentation, including VNet \cite{milletari2016v}, SegResNet \cite{myronenko20183d}, and nnUNet  \cite{isensee2018nnu}. \new{VNet  is a variant of UNet that introduces  short residual connections at each stage. Compared with UNet, SegResNet  contains an additional variational autoencoder branch. Finally,  nnUNet  is more flexible than UNet in three aspects: (1) residual connection in convolution blocks, (2) anisotropic kernel sizes and strides in each layer, and (3) deep supervision heads.} Deep learning has been applied to  lymphoma segmentation,  yielding promising results. In \cite{li2019densex}, Li et al. proposed a DenseX-Net-based lymphoma segmentation model with a two-flow architecture for 3D PET-CT images: a segmentation flow (DenseU-Net) for lymphoma segmentation  and a reconstruction flow (encoder-decoder)  for learning semantic representation of different lymphomas. In \cite{hu2020lymphoma}, Hu et al. introduced a multi-source fusion model for lymphoma segmentation with PET images. First, three 2D and one 3D segmentation models were trained with three orthogonal views and one 3D image, respectively. The four  segmentation maps were then fused by a convolutional layer to get a final result. In \cite{blanc2020fully}, Blanc-Durand et al. proposed a nnUNet-based lymphoma segmentation network with additional validation of total metabolic tumor volume for 3D PET-CT images. In \cite{huang21a}, Huang et al. proposed to fuse the outputs of two UNets trained on CT and PET data, using Dempster's rule of combination \cite{shafer1976mathematical}, a combination operator of Dempster-Shafer theory (DST) (see Section \ref{sec:background} below). However, the outputs of the UNets were probabilities and this approach  did not harness the full power of DST.

\begin{figure}
\includegraphics[width=\textwidth]{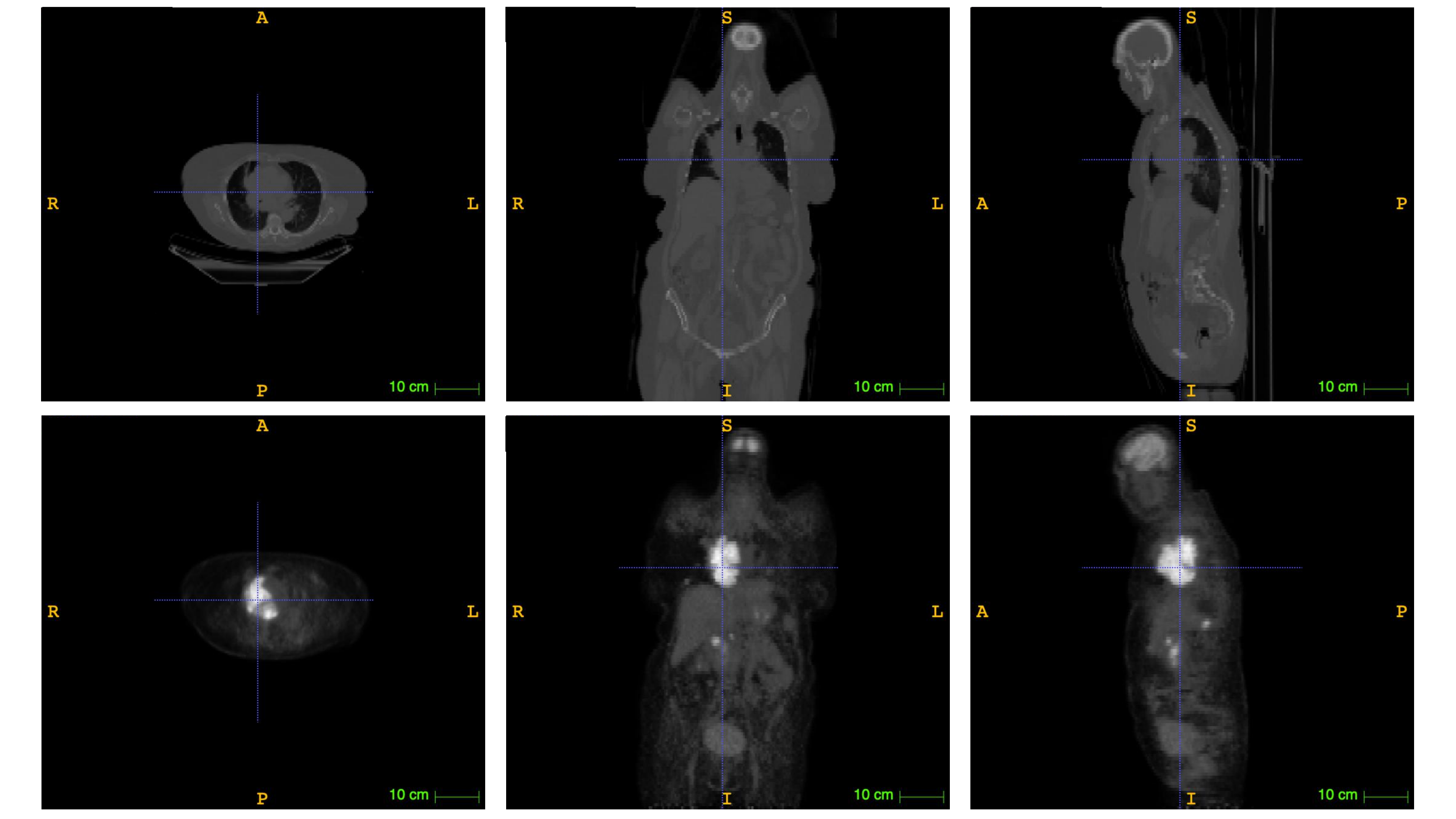}
\caption{Example of a patient with lymphomas. The first and second rows show, respectively, PET and CT slices for one patient in axial, sagittal and coronal views.}
\label{fig2}
\end{figure}

\paragraph{Uncertainty} In spite of the excellent performance of deep learning methods, the issue of quantifying  prediction uncertainty remains \cite{hullermeier2021aleatoric}. This uncertainty can be classified into three types: distribution, model, and data uncertainty. Distribution uncertainty is caused by training-test distribution mismatch (dataset shift) \cite{quinonero2009dataset}. Model uncertainty arises from limited training set size and model misspecification  \cite{mehta2019propagating}\cite{maddox2019simple}\cite{yu2019uncertainty}. Finally, sources of data uncertainty include  class overlap, label noise, and homo or hetero-scedastic noise \cite{ghesu2021quantifying}. Because of the limitations of medical imaging and labeling technology, as well as the need to use a large nonlinear parametric segmentation model, PET-CT image segmentation results are particularly tainted with uncertainty, which  limits the reliability of the segmentation. Figure \ref{fig2} shows  examples of PET and CT image slices for one patient with lymphomas. As can be seen, lymphomas in PET images  usually correspond to the brightest pixels, but organs such as the brain and bladder are also located in bright pixel areas, which may result in segmentation errors. Moreover,   lymphoma boundaries are blurred, which makes it hard to delineate lymphomas precisely.

\paragraph{Approaches to uncertainty modeling} Early approaches to  uncertainty quantification in machine learning  were based on Bayesian  theory \cite{hinton1993keeping}\cite{mackay1992practical}. The popularity of deep learning models has revived research of model uncertainty estimation and has given rise to specific methods such as  variational dropout \cite{gal2016dropout}\cite{tran2018bayesian}. 
In this paper, we explore a different approach based on DST \cite{dempster1967upper}\cite{shafer1976mathematical} \cite{denoeux20b}, a theoretical framework for  reasoning with imperfect (uncertain, imprecise, partial) information. DST was first introduced by Dempster \cite{dempster1967upper} and Shafer \cite{shafer1976mathematical} and was further popularized and developed by Smets \cite{smets1990combination}. Applications in machine learning were first introduced by Den{\oe}ux \cite{denoeux1995k,denoeux2000neural,denoeux2004evclus}. DST is based on the representation of elementary items of evidence by belief functions, and their combination by a specific operator called Dempster's rule of combination. In recent years, DST has generated considerable interest and has had great success in various fields, including information fusion \cite{pichon16}\cite{pichon19}\cite{chen21},  classification \cite{denoeux19f}\cite{gong21}\cite{imoussaten22}, clustering \cite{denoeux20b}\cite{denoeux21b}\cite{antoine21}, and image segmentation \cite{lian2018joint}\cite{huang2021belief}\cite{tong2021evidential}.

In this paper\footnote{This paper is an extended version of the short paper presented at the 6th International Conference on Belief Functions (BELIEF 2021) \cite{huang21b}.}, we propose a 3D PET-CT diffuse large B-cell lymphoma segmentation model based on DST and deep learning, which not only focuses on lymphoma segmentation accuracy but also on  uncertainty quantification using belief functions. The proposed segmentation model is composed of a UNet module for feature extraction and an evidential segmentation module for  uncertainty quantification and  decision-making. End-to-end learning is performed by minimizing the Dice loss function.  

The rest of the paper is organized as follows. The main concepts of DST are first recalled in Section \ref{sec:background}, and two approaches for computing belief functions in classification tasks are described in Section \ref{sec:evclass}. The proposed model is then introduced in Section \ref{sec:model}, and experimental results are reported in Section \ref{sec:exper}. Finally, Section \ref{sec:conc} concludes the paper.

\section{Dempster-Shafer theory}
\label{sec:background}

In this section, we  first recall some necessary notations and definitions regarding DST. Let $\Omega =\{\omega _{1},\omega _{2}, \ldots, \omega_{K}\} $ be a finite set of all possible answers some question, called the \emph{frame of discernment}. Evidence about the question of interest can be represented by a \emph{mass function} $m$, defined as a mapping from the power set $2^{\Omega}$ to $[0, 1]$ such that
\begin{equation}
    \sum _{A\subseteq \Omega }m(A)=1
    \label{eq:1}
\end{equation}
and $m(\emptyset)=0$, \new{where $\emptyset$ denotes the empty set}. Subsets $A \subseteq \Omega$ such $m(A)>0$ are called the \emph{focal sets} of $m$. Each mass $m(A)$ represents  a  share  of  a  unit  mass  of  belief  allocated  to focal set $A$, and  which  cannot  be  allocated  to  any  strict  subset  of $A$. The mass $m(\Omega)$ allocated to the whole frame can be seen as a degree of ignorance. Full ignorance is represented by the \emph{vacuous} mass function $m_?$ verifying $m_?(\Omega)=1$. A mass function is said to be \emph{Bayesian} if its focal sets are singletons, and \emph{logical} if it has only one focal set.

\paragraph{Discounting} Let $m$ be a mass function on $\Omega$ and $s$ a coefficient in $[0,1]$. The \emph{discounting} operation \cite{shafer1976mathematical} with discount rate $1-s$ transforms $m$ into a weaker, less informative mass function defined as follows:
\begin{equation}
\label{eq:disc}
^s m=s \, m +(1-s) \,m_?.
\end{equation}
As shown in \cite{smets94a}, coefficient $s$ can be interpreted as a degree of belief that the source of information providing mass function $m$ is reliable.

\paragraph{Simple mass functions} A mass function $m$ is said to be \emph{simple} if it can be obtained by discounting a logical mass function; it thus has the following form:
\begin{equation}
\label{eq:simple}
m(A)=s, \quad m(\Omega)=1-s,
\end{equation}
for some $A\subset \Omega$ such that $A\neq \emptyset$ and some $s\in [0,1]$, called the \emph{degree of support} in $A$. The quantity $w=-\ln(1-s)$ is called the \emph{weight of  evidence} associated to $m$ \cite[page 77]{shafer1976mathematical}. In the following, a simple mass function with focal set $A$ and weight of evidence $w$ will be denoted as $A^w$.

\paragraph{Belief and plausibility} Given a mass function $m$, \emph{belief} and \emph{plausibility} functions are defined, respectively, as follows:
\begin{equation}
   Bel(A) = \sum _{ B\subseteq A}m(B)
   \label{eq:3}
\end{equation}
and 
\begin{equation}
   Pl(A) = \sum _{B\cap A\neq \emptyset }m(B)=1-Bel(A^c),
   \label{eq:4}
\end{equation}
for all $A\subseteq \Omega$, where $A^c$ denotes the complement of $A$. The quantity $Bel(A)$ can be interpreted as a degree of support for $A$, while $Pl(A)$ can be interpreted as a measure of lack of support for the complement of $A$. 

\paragraph{Dempster's rule}  Two mass functions $m_{1}$ and $m_{2}$ derived from two independent items of evidence can be combined by considering each pair of a focal set $B$ of $m_1$ and a focal set $C$ of $m_2$, and assigning the product $m_1(B)m_2(C)$ to the intersection $B\cap C$. A normalization step is then necessary to ensure that the mass of the empty set is equal to zero. This operation, called \emph{Dempster's rule of combination} \cite{shafer1976mathematical} and denoted as $\oplus$, is formally defined   by $(m_{1}\oplus m_{2})(\emptyset)=0$ and
\begin{equation}
    (m_{1}\oplus m_{2})(A)=\frac{1}{1-\kappa }\sum _{B\cap C=A}m_{1}(B)m_{2}(C),
    \label{eq:5}
\end{equation}
for all $A\subseteq \Omega, A\neq \emptyset$, where $\kappa$ represents the \emph{degree of conflict} between $m_{1}$ and $m_{2}$ equal to
\begin{equation}
    \kappa=\sum _{B\cap C=\emptyset}m_{1}(B)m_{2}(C).
    \label{eq:6}
\end{equation}
The combined mass $m_{1}\oplus m_{2}$ is called the \emph{orthogonal sum} of $m_1$ and $m_2$.
It can easily be checked that the orthogonal sum of two simple mass functions $A^{w_1}$ and $A^{w_2}$ with the same focal set $A$ is the simple mass function $A^{w_1+w_2}$: Dempster's rule thus adds up weights of evidence.

\paragraph{Decision-making} After aggregating all the available evidence in the form of a mass function, it is often necessary to make a final decision. Decision-making based on belief functions for classification tasks has been studied in \cite{denoeux97}, and, more recently, by Ma and Den{\oe}ux in \cite{ma21}. The reader is referred to Ref. \cite{denoeux2019decision} for a recent review of decision methods based on belief functions. 
Here, we briefly introduce the approach used in this paper. 
Consider a classification task with $K$ classes in the set $\Omega=\{\omega_1,\ldots,\omega_K\}$. Assume that the utility of selecting the correct class is 1, and the utility of an error is 0. As shown in \cite{denoeux97}, the lower and upper expected utilities of selecting class $\omega_k$ are then, respectively, $Bel(\{\omega_k\})$ and  $Pl(\{\omega_k\})$. A pessimistic decision-maker (DM)  maximizing the lower expected utility will then select the class with the highest degree of belief, while an optimistic DM minimizing the upper expected utility will  select the most plausible class. Alternatively, the Hurwicz criterion consists in maximizing a weighted sum of the lower and upper expected utility. In the decision context, we then select the class $\omega_k$ such that $(1-\xi) Bel(\{\omega_k\})+ \xi Pl(\{\omega_k\})$ is maximum, where $\xi$ is an optimism index. Another approach, advocated by Smets in the Transferable Belief Model \cite{smets94a}, is to base decisions on the pignistic probability distribution, defined as
\begin{equation}
    p_m(\omega)=\sum_{\{A\subseteq \Omega: \omega\in A\}} \frac{m(A)}{|A|}
\end{equation}
for all $\omega \in \Omega$.

\section{Evidential classifiers}
\label{sec:evclass}

In this section, we review two methods for designing classifiers that output mass functions, referred to as \emph{evidential classifiers}. The evidential neural network (ENN) classifier introduced in \cite{denoeux2000neural} is first  recalled in Section \ref{subsec:enn}. A new model based on the interpretation of a  radial basis function (RBF) network as combining of simple mass functions by Dempster's rule, inspired by \cite{denoeux19d}, is then  described in Section \ref{subsec:nnweights}. The two models are compared experimentally in Section \ref{subsec:compar}.

\subsection{Evidential neural network}
\label{subsec:enn}

In \cite{denoeux2000neural}, Den{\oe}ux proposed the ENN classifier, in which mass functions are computed based on distances to prototypes. The basic idea is to consider each prototype as a piece of evidence, which is discounted based on its distance to the input vector. The evidence from different prototypes is then pooled by Dempster's rule \eqref{eq:5}. We provide a brief introduction to the ENN model in this section. 

\begin{figure}
\centering
\includegraphics[width=0.8\textwidth]{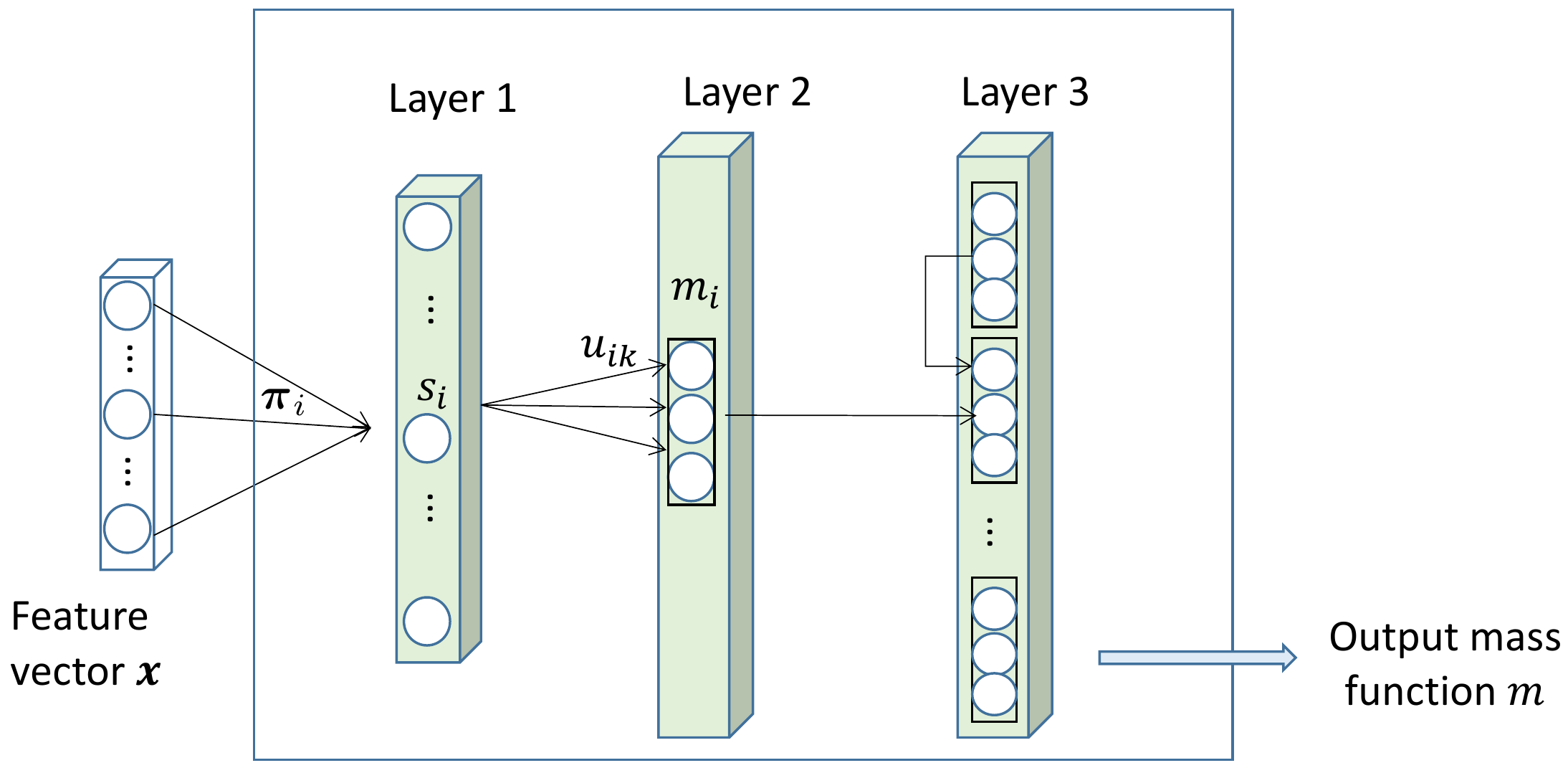}
\caption{Evidential neural network.}
\label{fig:ENN}
\end{figure}

The ENN classifier is composed on an input layer of $H$ neurons (where $H$ is the dimension of input space), two hidden layers and an output layer (Figure \ref{fig:ENN}). The first input layer is composed of $I$ units, whose weights vectors are prototypes \new{$\bp_1,\ldots, \bp_I$} in input space. The activation of unit $i$ in the prototype layer is
\begin{equation}
    s_i=\alpha _i \exp(-\gamma_i d_i^2),   
    \label{eq:8}
\end{equation}
where $d_i=\left \| \bx-\bp_i \right \| $ is the Euclidean distance between input vector $\bx$ and prototype $\bp_i$, $\gamma_i>0$ is a scale parameter,  and $\alpha_i \in [0,1]$ is an additional parameter. 

The second hidden layer computes mass functions $m_i$ representing the evidence of each prototype $\bp_i$, using the following equations:  
\begin{subequations}
\begin{align}
m_i(\{\omega _{k}\})&=u_{ik}s_i, \quad k=1,..., K\\
m_{i}(\Omega)&=1-s_i, 
\label{eq:9}
\end{align}
\end{subequations}
where $u_{ik}$ is the membership degree of prototype $i$ to class $\omega_k$, and $\sum _{k=1}^K u_{ik}=1$. Mass function $m_i$ can thus be seen as a discounted Bayesian mass function, with discount rate $1-s_i$; its focal sets are singletons and $\Omega$. The mass assigned to $\Omega$ increases with the distance between $\bx$ and $\bp_i$. Finally, the third layer combines the $I$ mass functions $m_1,\ldots,m_I$ using Dempster's rule \eqref{eq:5}. The output mass function $m=\bigoplus_{i=1}^I m_i$ is a discounted Bayesian mass function that summarizes the evidence of the $I$ prototypes. Because the focal sets of $m$ are singletons and $\Omega$, the class with the highest degree of belief also has the highest plausibility and pignistic probability: consequently, the decision rules recalled in Section \ref{sec:background} are equivalent in this case. 

Let $\btheta$ denote the vector of all network parameters, composed of the $I$ prototypes $\bp_i$, their parameters $\gamma_i$ and $\alpha_i$, and their membership degrees $u_{ik}$, $k=1,\ldots,K$. In \cite{denoeux2000neural}, it was proposed to learn these parameters  by minimizing the regularized sum-of-squares loss function
\begin{equation}
 L_{SS}(\btheta)=\sum_{n=1}^{N}\sum_{k=1}^{K}(p_{nk}-y_{nk})^{2}+ \lambda  \sum_{i=1}^{I}\alpha_{i},  
 \label{eq:lossENN}
\end{equation}
where $p_{nk}$ is the pignistic probability of class $\omega_k$ for instance $n$, \new{$N$ is the number of training instances}, and $y_{nk}=1$ if the true class of instance $n$ is $\omega_k$, and  $y_{nk}=0$ otherwise.  The second term on the right-hand side of \eqref{eq:lossENN} is a regularization term, and $\lambda$ is hyperparameter that can be tuned by cross-validation.

The idea of applying the above model to features extracted by a convolutional neural network (CNN) was first proposed by Tong et al. in \cite{tong21b}. In this approach, the ENN module becomes a ``evidential layer'', which is plugged into the output of a CNN instead of the usual softmax layer. The feature extraction and evidential modules are trained simultaneously. A similar approach was applied in  \cite{tong2021evidential} to semantic segmentation.  In the next section, we present an alternative approach based on a radial basis function (RBF) network and weights of evidence.

\subsection{Radial basis function network}
\label{subsec:nnweights}

As shown in \cite{denoeux19d}, the calculations performed in the softmax layer of a feedforward neural network can be interpreted in terms of combination of evidence by Dempster's rule. The output class probabilities can be seen as normalized plausibilities according to an underlying belief function. Applying these ideas to a radial basis function (RBF) network, it is possible to derive an alternative evidential classifier with  properties similar to those of the ENN model recalled in Section  \ref{subsec:enn}.

Consider an RBF network with $I$ prototype (hidden) units. The activation of hidden unit $i$ is
\begin{equation}
    s_i=\exp(-\gamma_i d_i^2),   
    \label{eq:activRBF}
\end{equation}
where, as before, $d_i=\left \| \bx-\bp_i \right \| $ is the Euclidean distance between input vector $\bx$ and prototype $\bp_i$, and $\gamma_i>0$ is a scale parameter. For the application considered in this paper, we only need to consider  the case of binary classification with $K=2$ and $\Omega=\{\omega_1,\omega_2\}$. (The case where $K>2$ is also analyzed in \cite{denoeux19d}). Let $v_{i}$ be the weight of the connection between hidden unit $i$ and the output unit, and let $w_i=s_i v_i$ be the product of the output of unit $i$ and weight $v_i$. The quantities $w_i$ can be interpreted as weights of evidence for class $\omega_1$ or $\omega_2$, depending on the sign of $v_i$:
\begin{itemize}
    \item If $v_{i}\ge 0$, $w_i$  a weight of evidence for class $\omega_1$;
    \item If $v_{i}<0$, $-w_i$ is a weight of evidence for class $\omega_2$.
\end{itemize}
To each prototype $i$  can, thus, be associated  the following simple mass function:
\[
m_{i}=\{\omega_1\}^{w_{i}^+} \oplus \somega{2}^{w_{i}^-},
\]
where \new{$w_{i}^+=\max(0,w_i)$ and $w_{i}^-=-\min(0,w_i)$} denote, respectively, the positive and negative parts of $w_i$. Combining the evidence of all prototypes in favor of $\omega_1$ or $\omega_2$ by Dempster's rule, we get the mass function
\begin{equation}
\label{eq:mRBF1}
m=\bigoplus_{i=1}^I m_{i}=\{\omega_1\}^{w^+} \oplus \somega{2}^{w^-},
\end{equation}
with $w^+=\sum_{i=1}^I w_{i}^+$ and $w^-=\sum_{i=1}^I w_{i}^-$. 
In \cite{denoeux19d}, the normalized plausibility of $\omega_1$ corresponding to mass function $m$ was shown to have the following expression:
\begin{equation}
\label{eq:p}
    p(\omega_1)=\frac{Pl(\somega{1})}{Pl(\somega{1})+Pl(\somega{2})}=\frac{1}{1+\exp(-\sum_{i=1}^I v_i s_i)},
\end{equation}
i.e., it is the output of a unit with a logistic activation function. When training an RBF network with a logistic output unit, we thus actually combine evidence from each of the prototypes, but the combined mass function remains latent. In \cite{denoeux19d}, mass function $m$ defined by \eqref{eq:mRBF1} was shown to have the following expression:
\begin{subequations}
\label{eq:m12}
\begin{align}
m(\somega{1}) &= \frac{[1-\exp(-w^+)]\exp(-w^-)}{1-\kappa} \\
m(\somega{2}) &= \frac{[1-\exp(-w^-)]\exp(-w^+)}{1-\kappa}\\
\label{eq:m12Omega}
m(\Omega) &= \frac{\exp(-w^+-w^-)}{1-\kappa}=\frac{\exp(-\sum_{i=1}^I|w_i|)}{1-\kappa},
\end{align}
where
\begin{equation}
\label{eq:conf12}
\kappa=[1-\exp(-w^+)] [1-\exp(-w^-)]
\end{equation}
\end{subequations}
is the degree of conflict between mass functions $\{\omega_1\}^{w^+}$ and $\{\omega_2\}^{w^-}$.

In the approach, we thus simply need to train a standard RBF network with $I$ prototype layers and one output unit with a logistic activation function, by minimizing a loss function such as, e.g., the regularized cross-entropy loss
\begin{equation}
\label{eq:lossRBF}
L_{CE}(\btheta)=-\sum_{n=1}^N \left(y_n\log p_n  + (1-y_n)\log(1-p_n)\right) + \lambda \sum_{i=1}^I w_i^2,
\end{equation}
where $p_n$ is the normalized plausibility of class $\omega_1$ computed from \eqref{eq:p} for instance $n$, \new{$y_n$ is class label of instance $n$ ($y_n=1$ if the true class of instance $n$ is $\omega_1$, and $y_n=0$ otherwise),} and $\lambda$ is a hyperparameter. We note that increasing $\lambda$ has the effect of decreasing the weights of evidence and, thus,  obtaining less informative mass functions.

\subsection{Comparison between the two models} 
\label{subsec:compar}

\begin{figure}
\begin{center}
\includegraphics[width=0.5\textwidth]{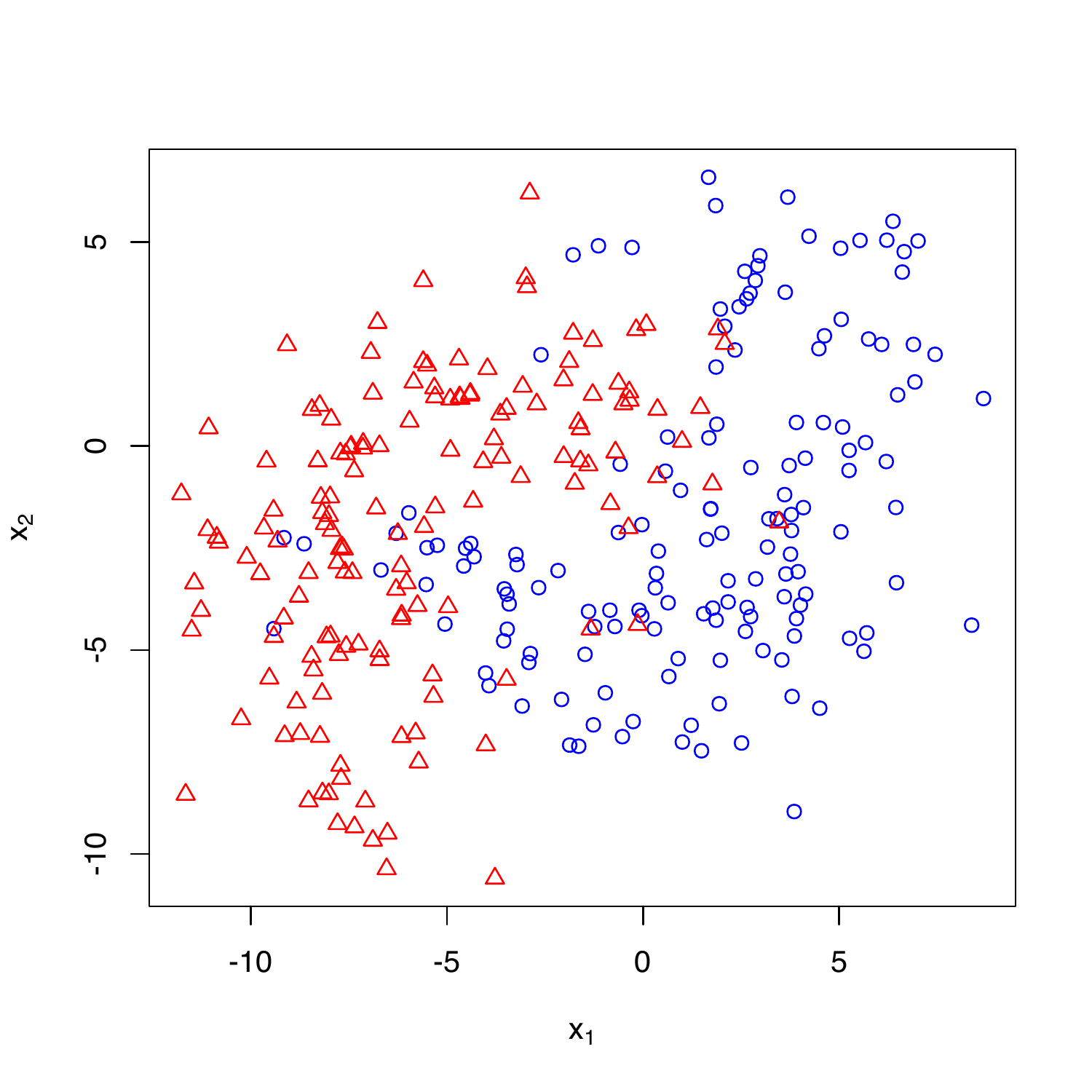}
\end{center}
\caption{Simulated data. \label{fig:bananas}}
\end{figure}

To compare the RBF model described in  Section \ref{subsec:nnweights} with the ENN model recalled in Section \ref{subsec:enn}, we consider the two-class dataset shown in Figure \ref{fig:bananas}. The two classes are randomly distributed around half circles with Gaussian noise and are separated by a nonlinear boundary. A learning set of size $N=300$ and a test set of size 1000 were generated from the same distribution. 

\begin{figure}
\subfloat[\label{fig:error}]{\includegraphics[width=0.5\textwidth]{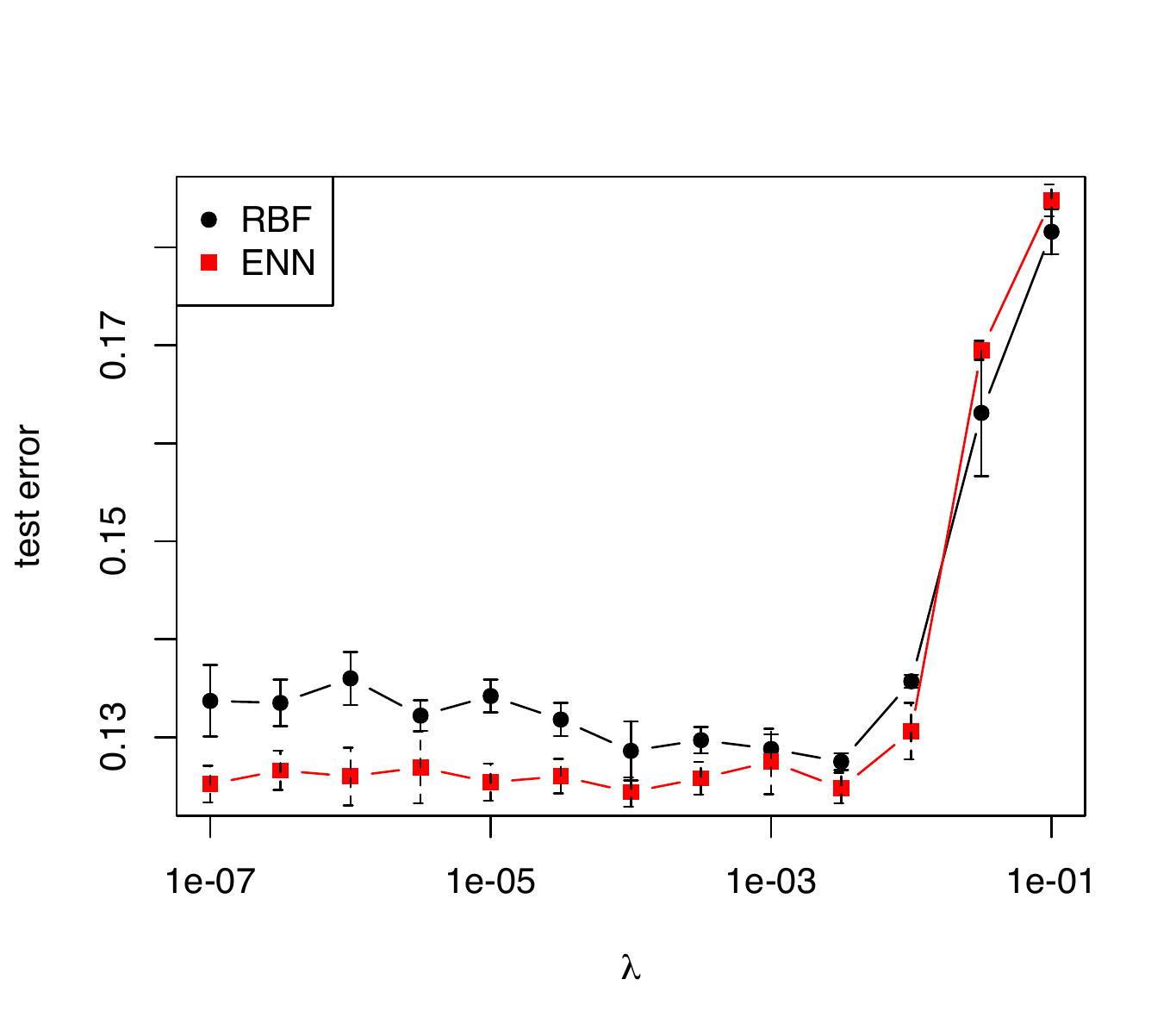}}
\subfloat[\label{fig:uncert}]{\includegraphics[width=0.5\textwidth]{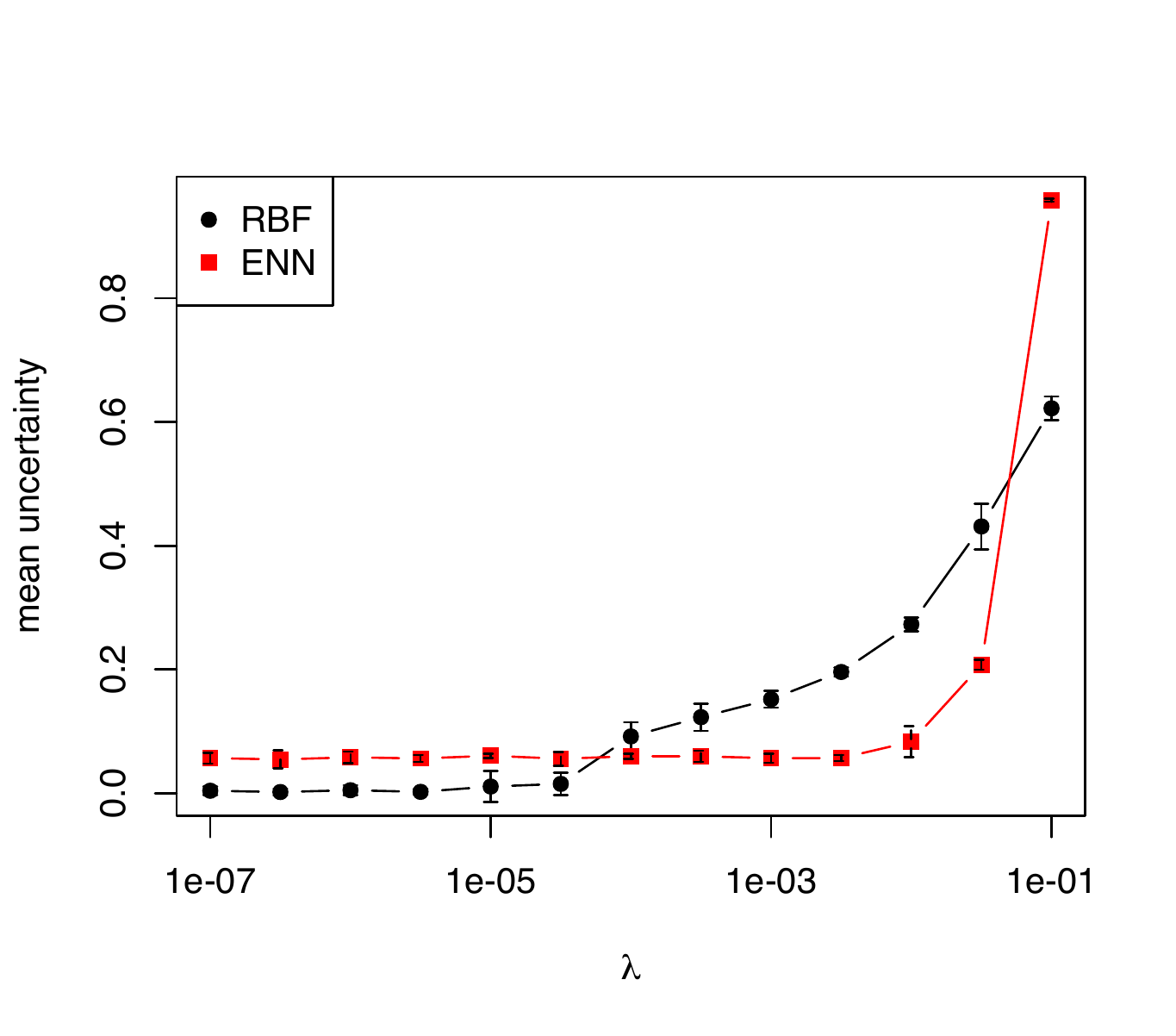}}
\caption{Test error rates (a) and mean uncertainty (b) for the ENN and RBF models, as functions of regularization parameter $\lambda$. \label{fig:compar_err}}
\end{figure}

An ENN and a RBF network were initialized with $I=6$ prototypes generated by the $k$-means algorithm and were trained on the learning data. Figures \ref{fig:error} and \ref{fig:uncert} show, respectively, the test error rate and the mean uncertainty (defined as the average mass assigned to the frame $\Omega$), as functions of hyperparameter $\lambda$ in \eqref{eq:lossENN} and \eqref{eq:lossRBF}, for 10 different runs of both algorithms with different initializations. As expected, uncertainty increases with $\lambda$ for both models, but the ENN model appears to be less sensitive to $\lambda$ as compared to the RBF model.  Both models achieve similar minimum error rates for $\lambda$ around $10^{-3}$, and have similar mean uncertainties for $\lambda=10^{-4}$. 

\begin{figure}
\centering
\subfloat[\label{fig:contourRBFomega1}]{\includegraphics[width=0.35\textwidth]{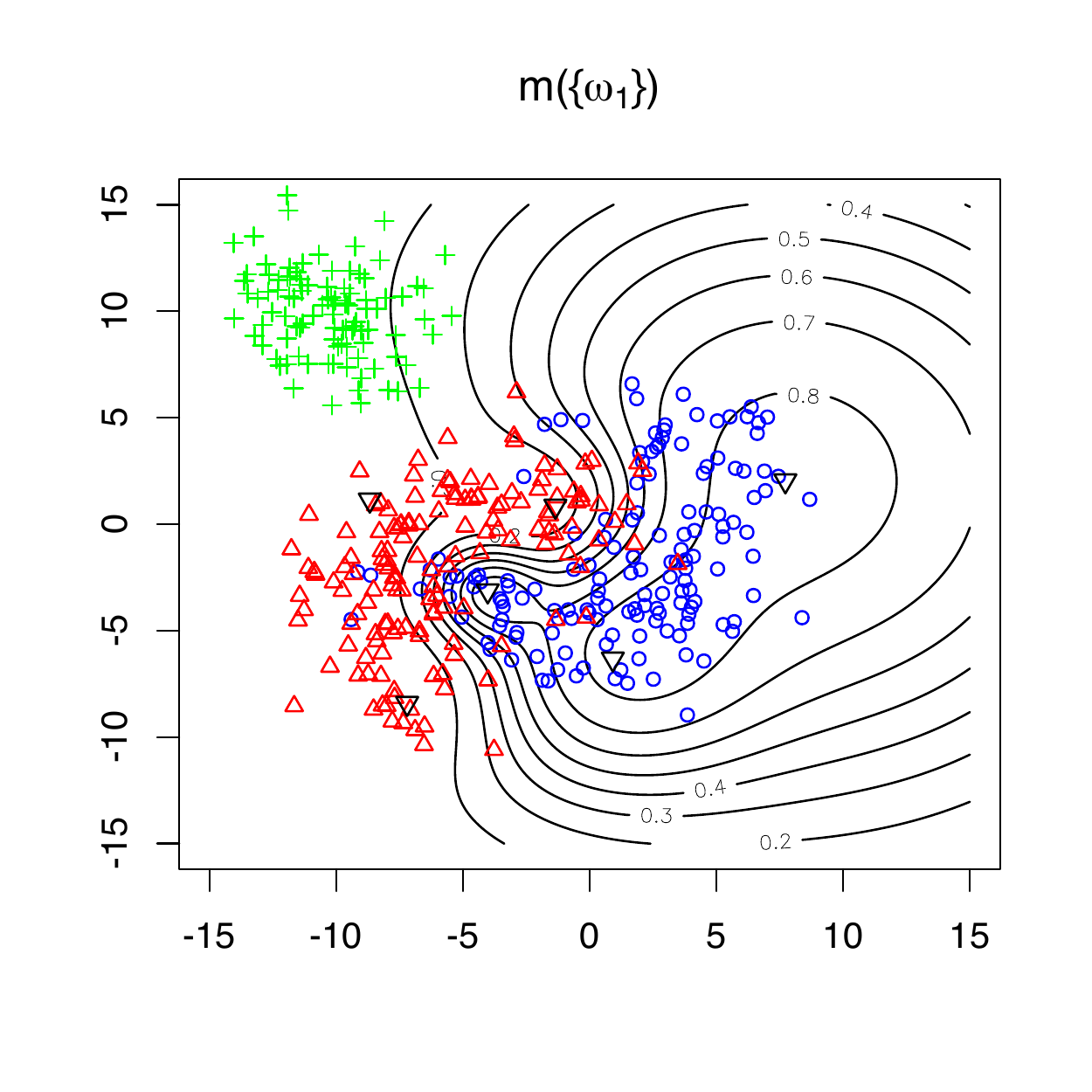}}
\subfloat[\label{fig:contourENNomega1}]{\includegraphics[width=0.35\textwidth]{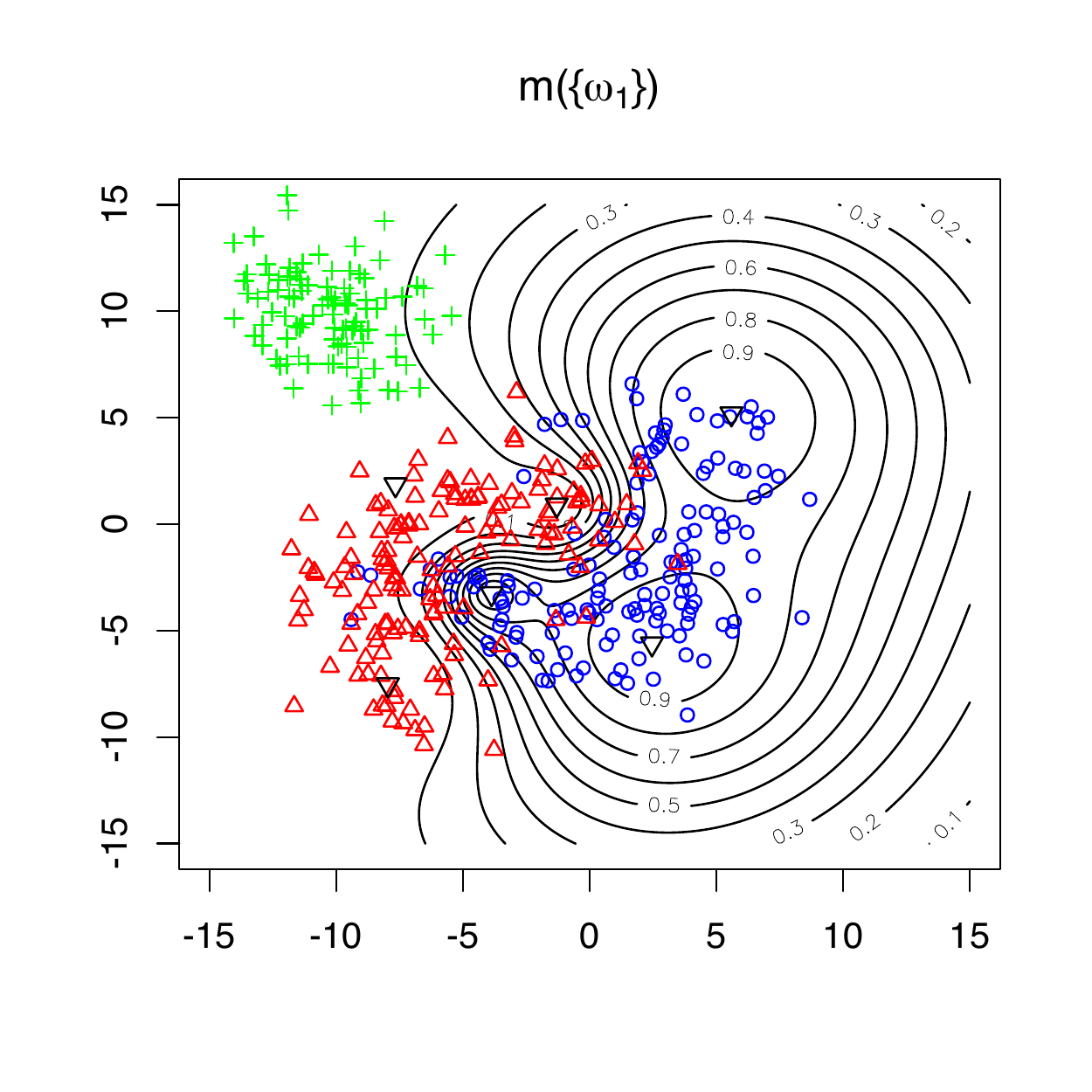}}\\
\subfloat[\label{fig:contourRBFomega2}]{\includegraphics[width=0.35\textwidth]{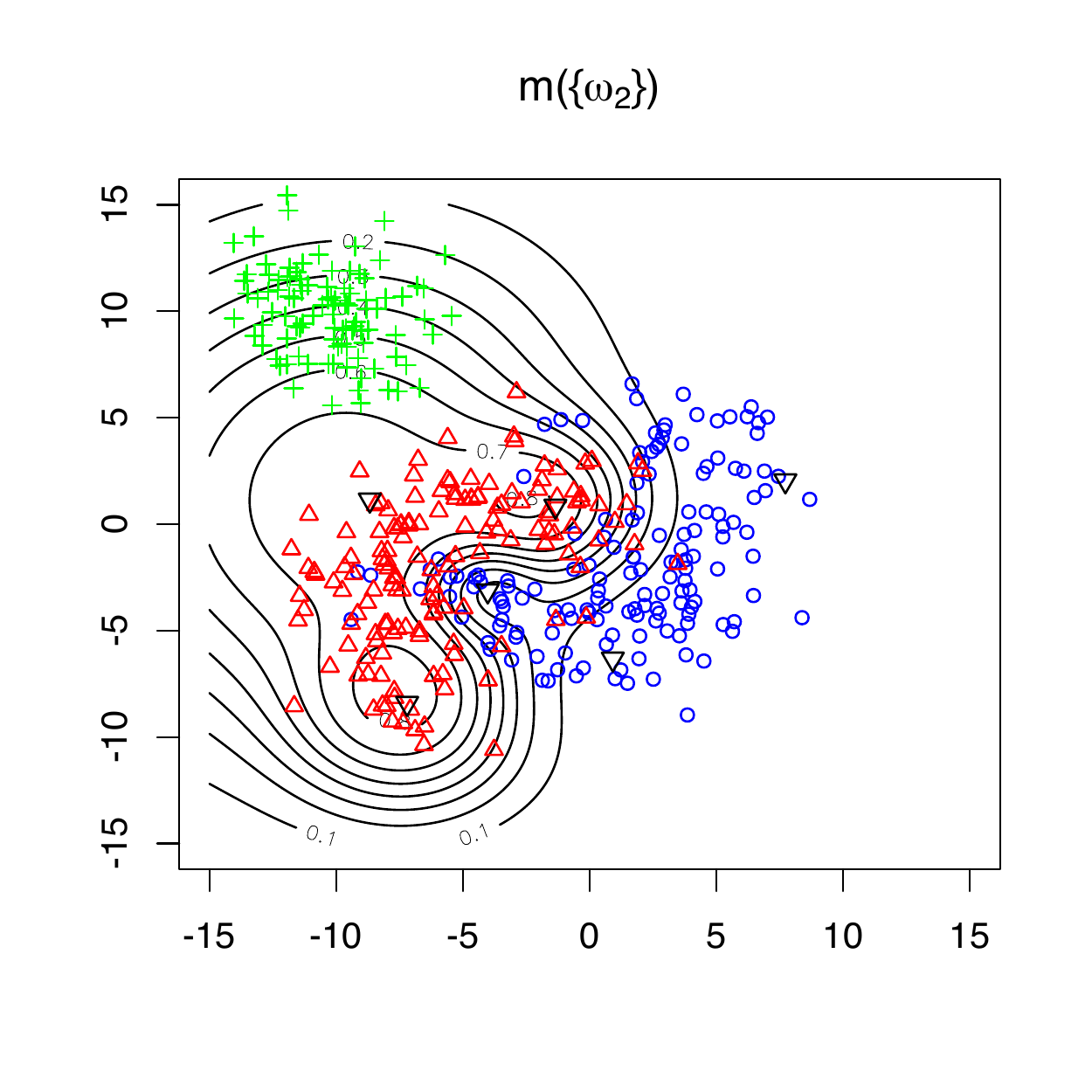}}
\subfloat[\label{fig:contourENNomega2}]{\includegraphics[width=0.35\textwidth]{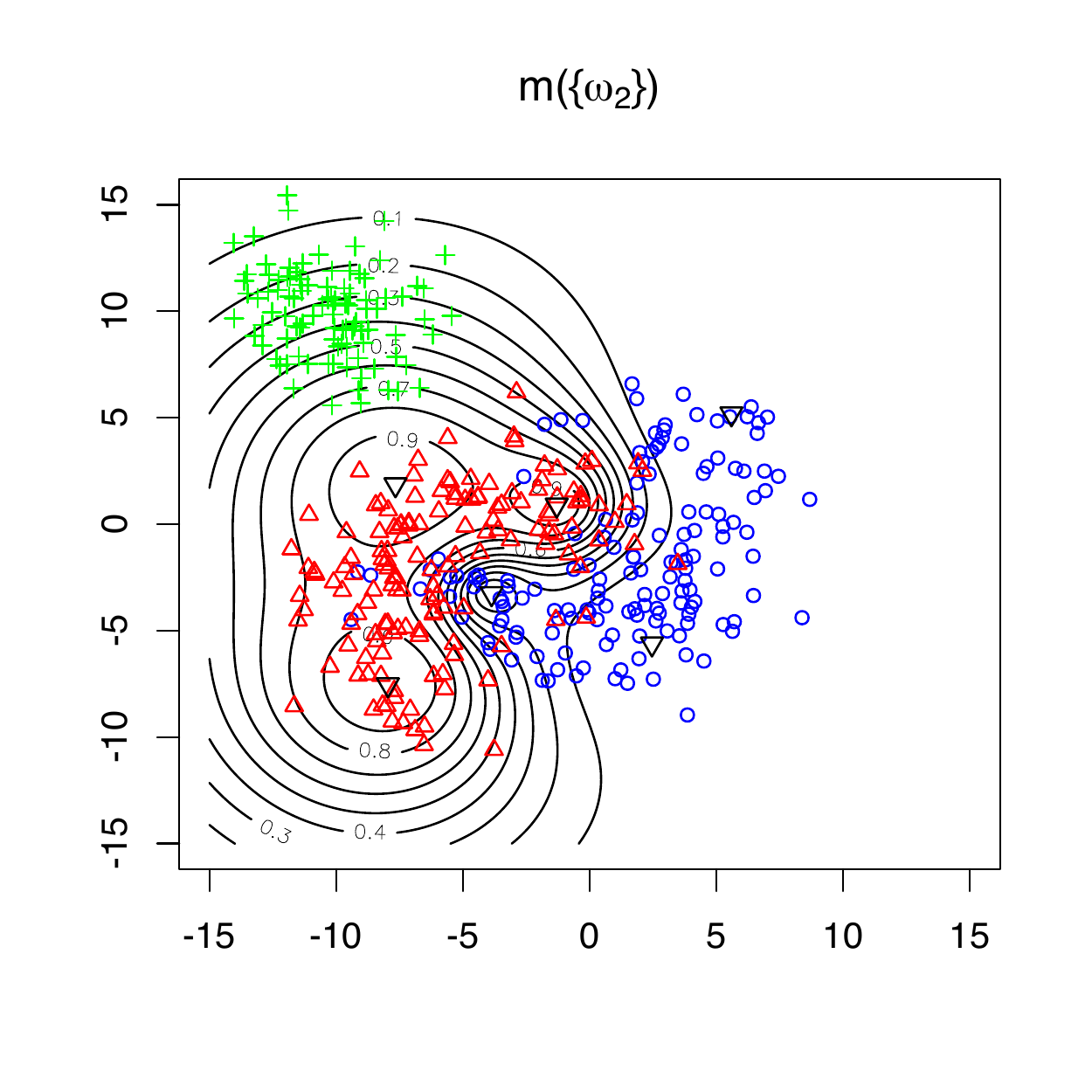}}\\
\subfloat[\label{fig:contourRBFOmega}]{\includegraphics[width=0.35\textwidth]{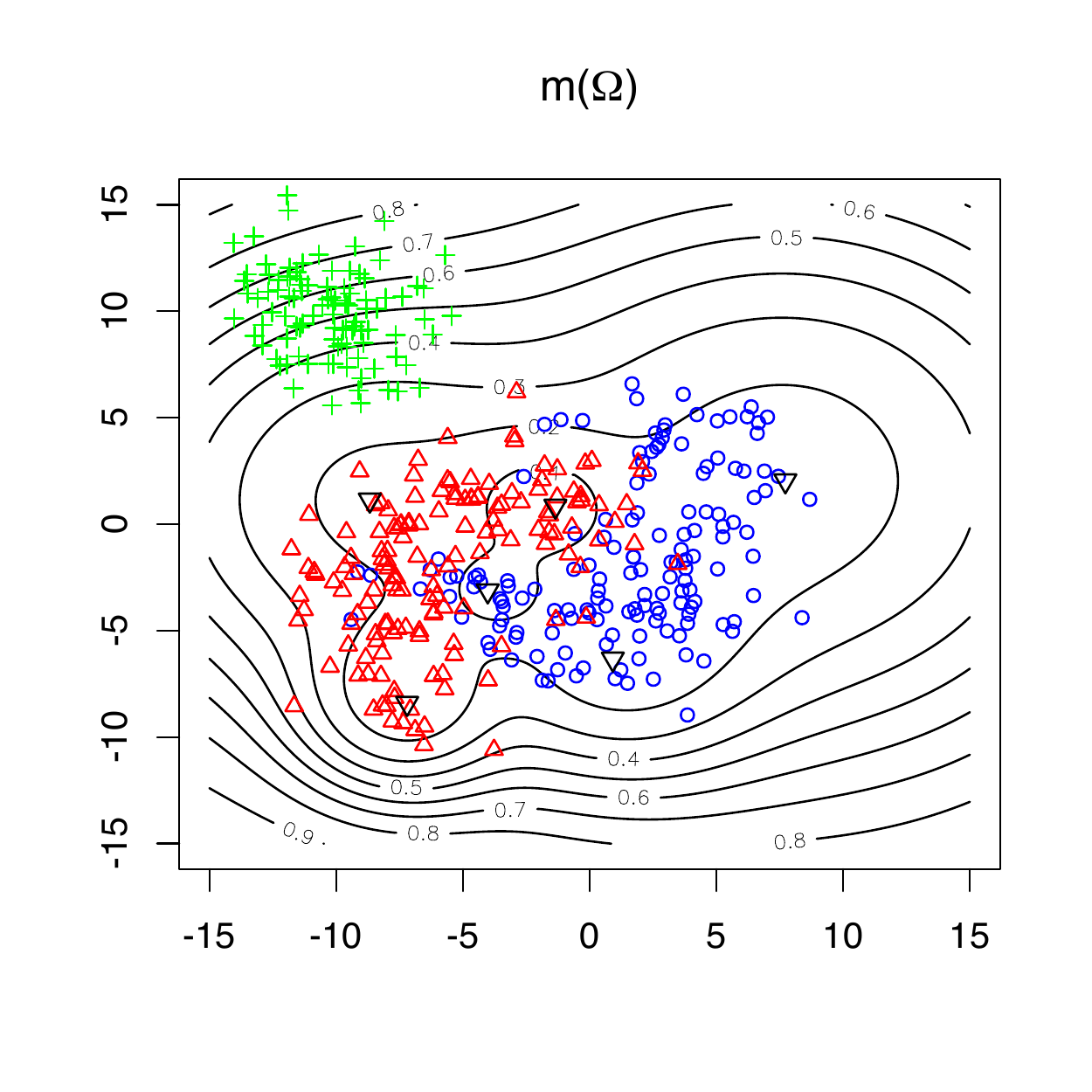}}
\subfloat[\label{fig:contourENNOmega}]{\includegraphics[width=0.35\textwidth]{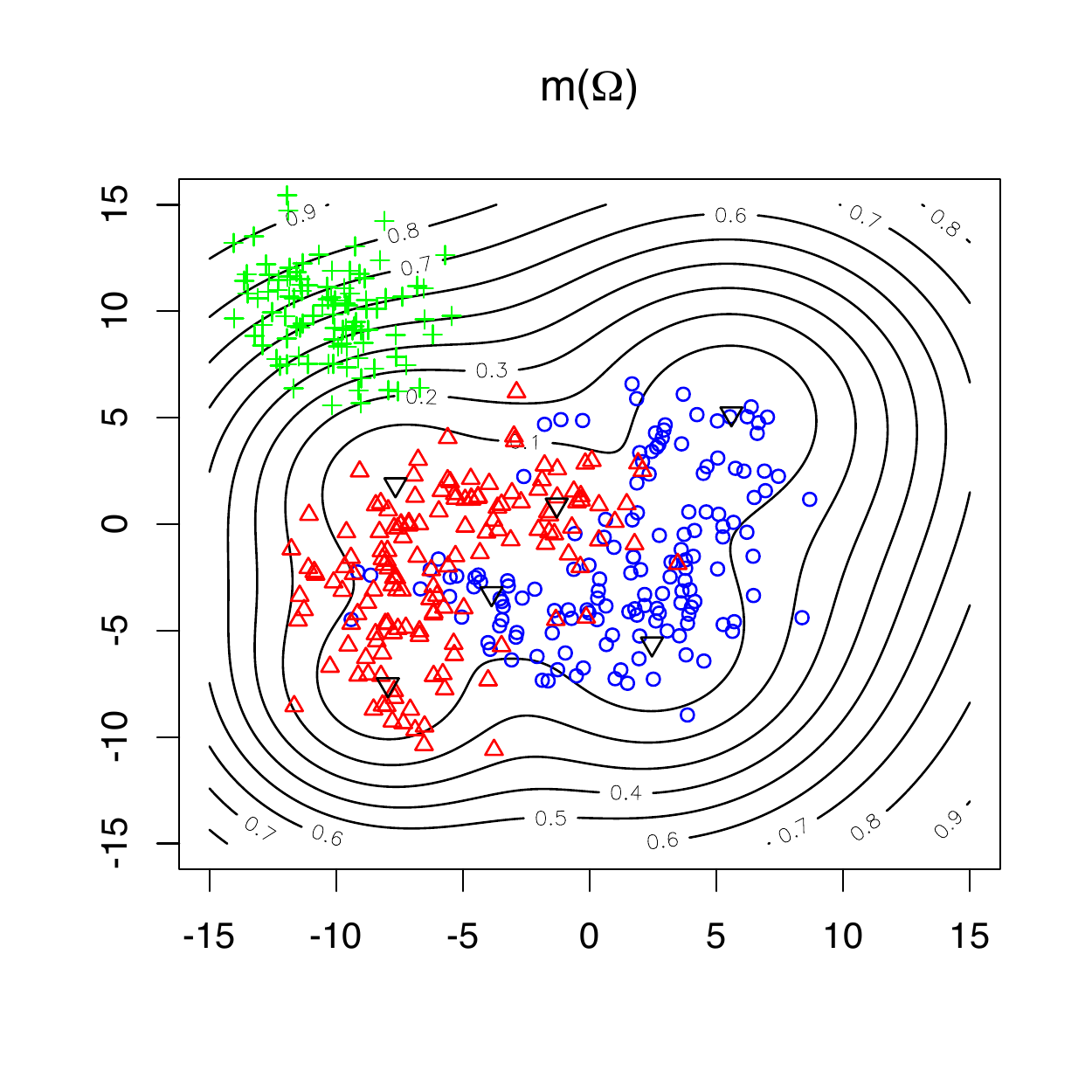}}
\caption{Contours of the mass assigned to $\{\omega_1\}$, $\{\omega_2\}$ and $\Omega$ by the RBF  (left column) and ENN (right column) models. The training data are displayed in blue and red, and the third class (absent from the training data) is shown in green. Training was done with $\lambda=0.001$ for the two models. 
\label{fig:contours}}
\end{figure}

\begin{figure}
\centering
\subfloat[\label{fig:omega1}]{\includegraphics[width=0.4\textwidth]{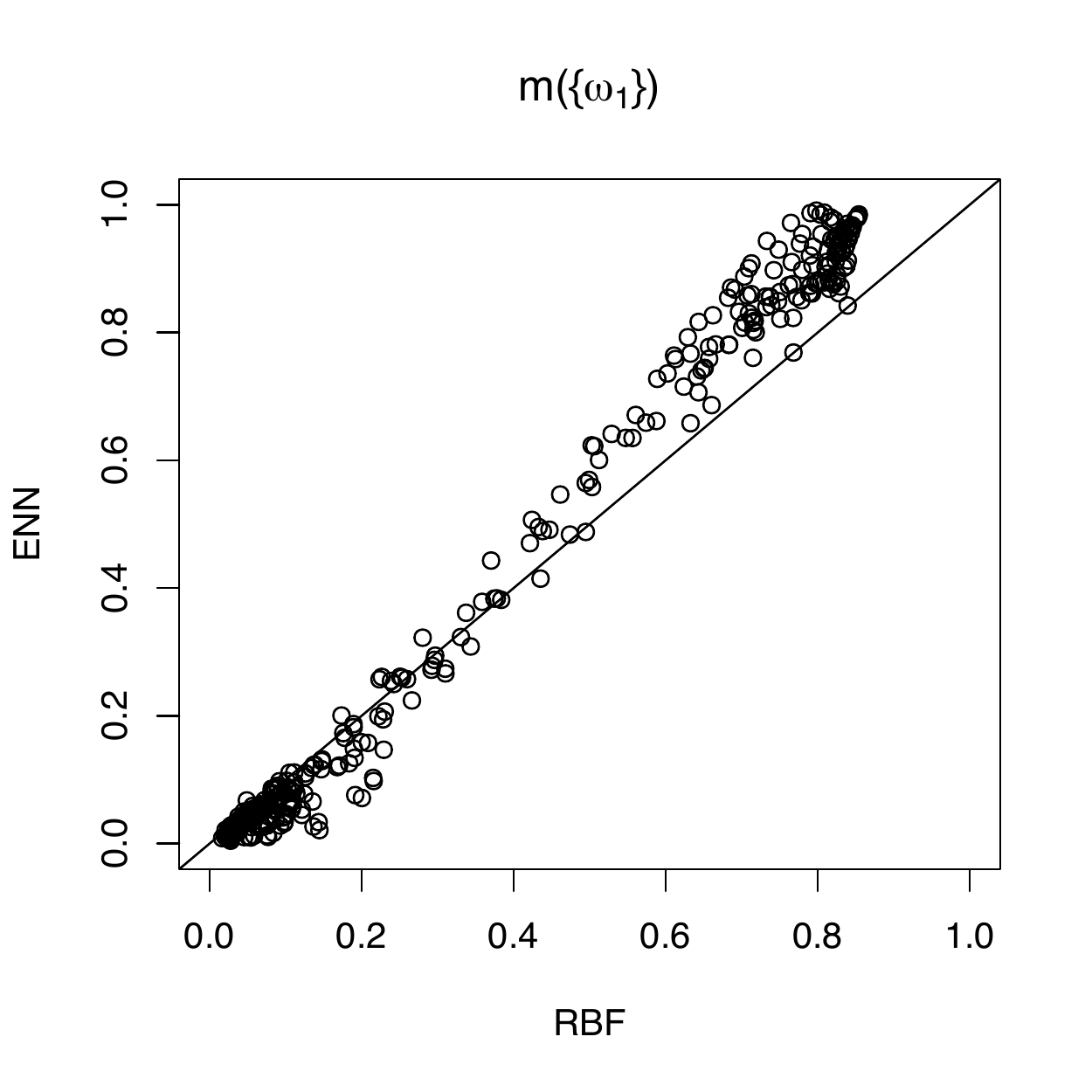}}
\subfloat[\label{fig:omega2}]{\includegraphics[width=0.4\textwidth]{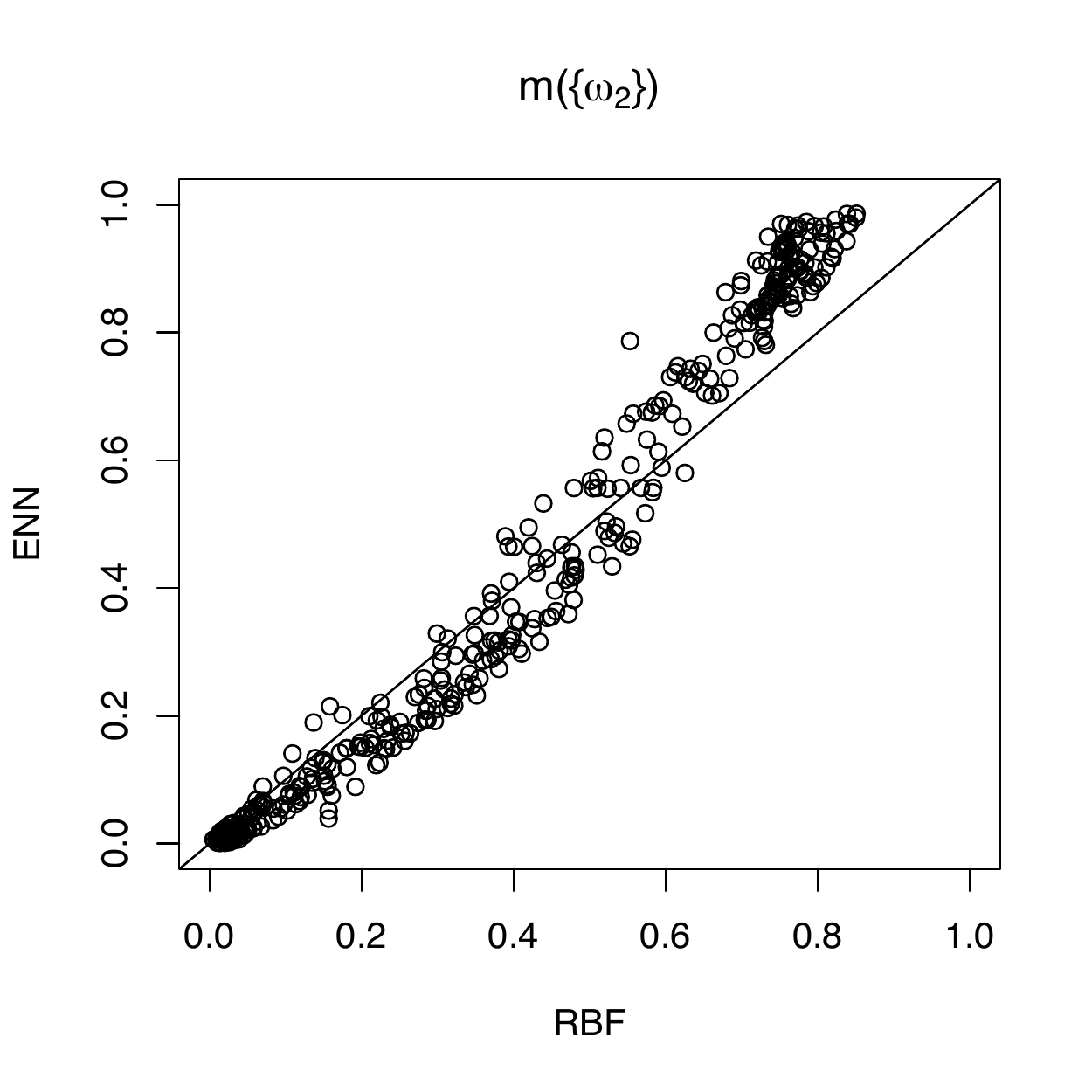}}\\
\subfloat[\label{fig:Omega}]{\includegraphics[width=0.4\textwidth]{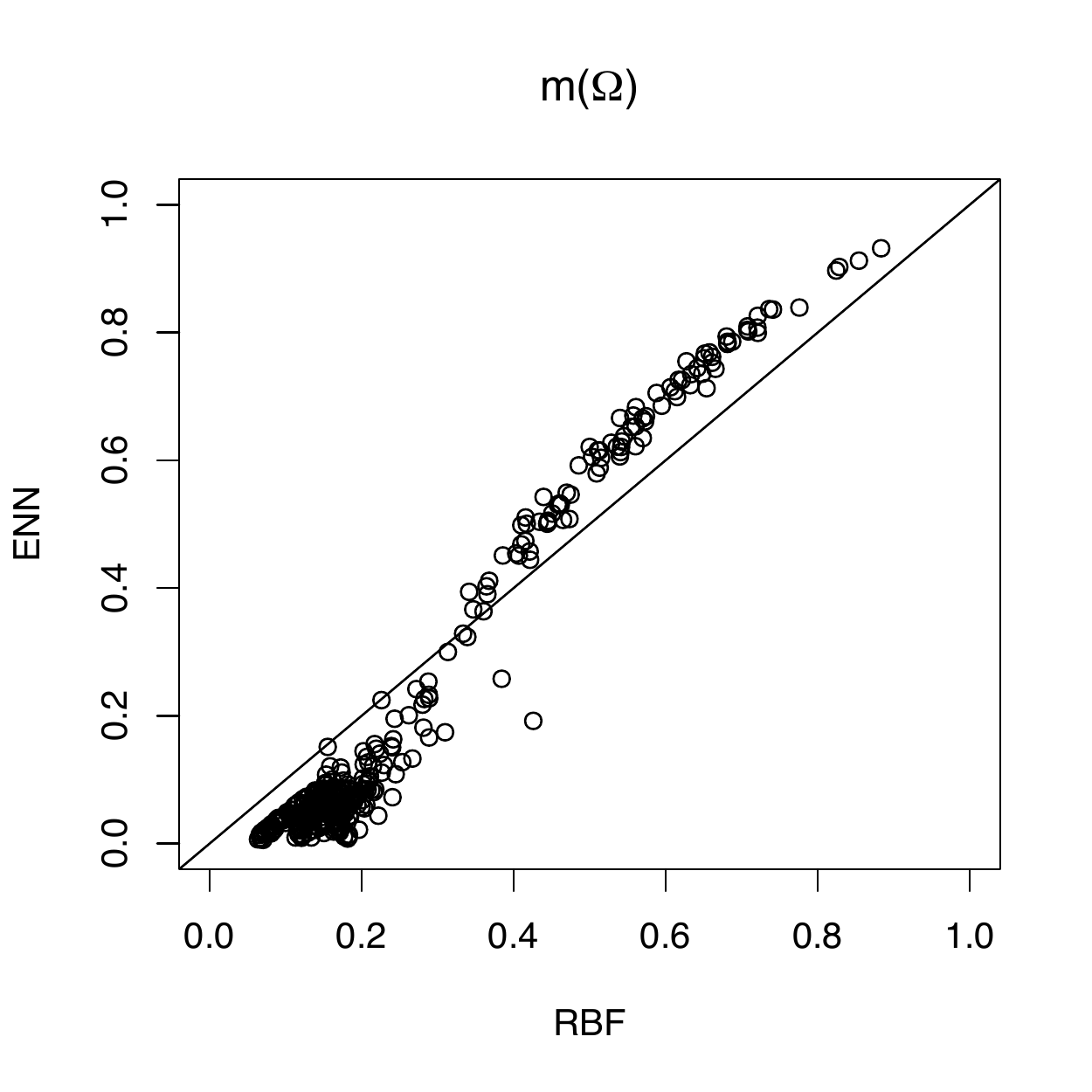}}
\caption{Masses computed by the RBF network (horizontal axis) versus the ENN model (vertical axis) for the extended dataset. \label{fig:compar_mass}}
\end{figure}

\new{As shown in \cite{denoeux2000neural}, the robustness of the ENN model arises from the fact that, when the input $\bx$ is far from all prototypes, the output mass function $m$ is close to the vacuous mass function. This property, in particular, makes the network capable of detecting observations generated from a distribution that is not represented in the learning set. From \eqref{eq:m12Omega}, we can expect the RBF network model to have a similar property: if $\bx$ is far from all prototypes, all weights of evidence $w_i$ will be small and the mass $m(\Omega)$ will be close to unity. To compare the mass functions computed by the two models, not only in regions of high density where training data are present, but also in regions of low density, we introduced  a third class in the test set, as shown in Figure \ref{fig:contours}}. Figure \ref{fig:compar_mass} shows scatter plots of masses on each of the focal sets computed for the two models trained with $\lambda=10^{-3}$ and applied to an extended dataset \new{composed of the learning data and the third class}. We can see that the mass functions are quite similar. Contour plots  shown in Figure \ref{fig:contours} confirm this similarity.

\section{Proposed model}
\label{sec:model}

\begin{figure}
\includegraphics[width=\textwidth]{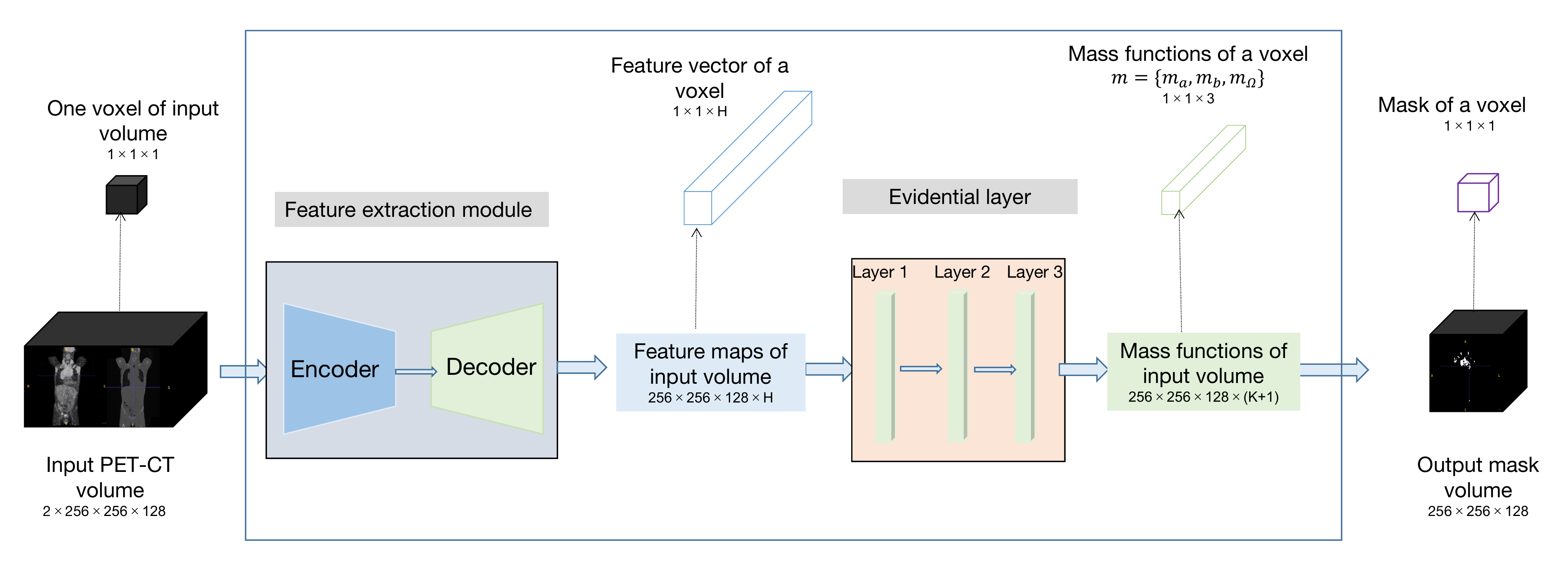}
\caption{Global lymphoma segmentation model.}
\label{fig:arch}
\end{figure}

 The main idea of this work is to hybridize a deep medical image segmentation model with one of the evidential classifiers introduced in Section \ref{sec:evclass}. Figure \ref{fig:arch} shows the global lymphoma segmentation architecture, composed of an encoder-decoder feature extraction module (UNet), and an evidential layer based one of the two models described in Section \ref{sec:evclass}. \new{The  input is the concatenated PET-CT image volume provided as  a tensor of size $2 \times256 \times256 \times 128$, where 2 corresponds to the number of modality channels, and $256 \times256 \times 128$ is the size of each input volume. The PET-CT image volumes are first fed into the feature extraction module, which outputs high-level features in the form of a  tensor of size $256 \times256 \times 128 \times H$, where $H$ is the number of features computed at each voxel. This tensor is then fed into the evidential layer, which  outputs mass functions representing evidence about the class of each  voxel, resulting in a tensor of size $256 \times256 \times 128 \times (K+1)$, where $K+1$ is the number of masses (one for each class and one for the frame of discernment $\Omega$).} The whole network is trained end-to-end by minimizing a regularized Dice loss. The different components of this model are described in greater detail below.

\begin{figure}
\includegraphics[width=\textwidth]{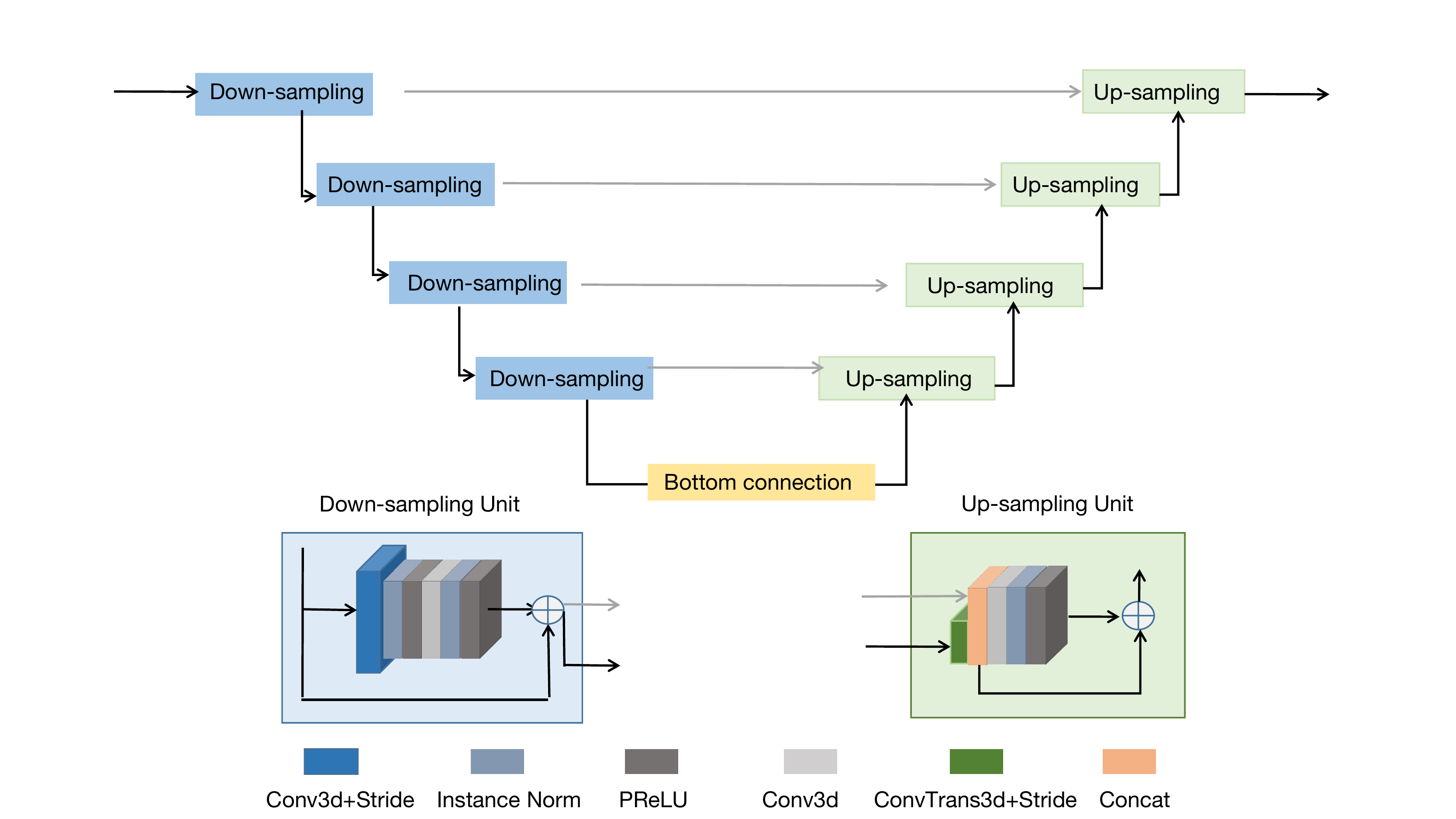}
\caption{Feature extraction module.}
\label{fig-unet}
\end{figure}

\paragraph{Feature extraction module} The feature extraction module is based on a UNet \cite{ronnebergerconvolutional} with residual encoder and decoder layers \cite{kerfoot2018left}, as shown in Figure~\ref{fig-unet}. Each down-sampling layer (marked in blue) is composed of convolution, normalization, dropout and activation blocks. Each up-sampling layer (marked in green) is composed of transpose convolution, normalization, dropout and activation blocks. The last layer (marked in yellow) is the bottom connection which does not down or up-sample the data. In the experiments reported in Section \ref{sec:exper}, the channels \new{(number of filters)} were set as $(8, 16, 32, 64, 128)$ with kernel size equal to 5 and \new{convolutional} strides equal to $(2, 2, 2, 2)$. The spatial dimension, input channel and output channel of the module were set, respectively, as 3, 2, \new{and the number $H$ of extracted features. (Experiments with several values of $H$ are reported in Section \ref{subsec:sensitivity}).} 
\new{The dropout rate was set as 0 and no padding operation was applied. Instance normalization \cite{ulyanov2017improved} was used to perform intensity normalization across the width, height and depth of a single feature map of a single example. The Parametric Rectified Linear Unit (PReLU) function \cite{he2015delving},  which generalizes the traditional rectified unit with a slope for negative values, was used as the activation function.} \new{For each input voxel, the feature extraction module outputs a $1 \times H$ feature vector, which is fed into the evidential layer.}

\paragraph{Evidential layer}
A probabilistic network with a softmax output layer may assign voxels a high probability of belonging to one class while the segmentation uncertainty is actually high because, e.g., the \new{feature vector describing that voxel is far away from feature vectors presented during training}. Here, we propose to plug-in one of the evidential classifiers described in Section \ref{sec:evclass} at the output of the feature extraction module. The ENN or RBF classifier then takes as inputs the high-level feature vectors computed by the UNet and computes, for each voxel $n$, a mass function $m_n$ on the frame $\Omega=\{\omega_1,\omega_2\}$, where $\omega_1$ and $\omega_2$ denote, respectively, the background and the lymphoma class. We will use the names ``ENN-UNet''  and ``RBF-UNet'' to designate the two variants of the architecture. 

\paragraph{Loss function}
The whole network is trained end-to-end by minimizing a regularized Dice loss. We use the Dice loss instead of the original cross-entropy loss in UNet because the quality of the segmentation is finally assessed by the Dice coefficient. The Dice loss is defined as
\begin{equation}
    \textsf{loss}_{D}=1-\frac{2 \sum_{n=1}^{N} S_n G_n}{ \sum_{n=1}^{N} S_n+ \sum_{n=1}^{N} G_n},
    \label{eq:Dice_loss}
\end{equation}
where $N$ is the number of voxels in the image volume, $S_n$ is the output pignistic probability of the tumor class (i.e., $m_n(\{\omega_2\})+ m_n(\Omega)/2$) for voxel $n$, and $G_n$ is ground truth for voxel $n$, defined as $G_n=1$ if voxel $n$ corresponds to a tumor, and $G_n=0$. The regularized loss function is 
    \begin{equation}
        \textsf{loss}=\textsf{loss}_{D} +\lambda R,  
    \label{eq:22}
    \end{equation}
where $\lambda$ is the regularization coefficient and $R$ is a regularizer defined either as $R=\sum_i \alpha_i$ if the ENN classifier is used in the ES module, or as $R=\sum_i v_i^2$ if the RBF classifier is used. The regularization term allows us to decrease the influence of unimportant prototypes and avoid overfitting. 


\section{Experiments}
\label{sec:exper}

The model introduced in Section \ref{sec:model} was applied to a set of PET-CT data recorded on patients with lymphomas\footnote{\new{Our code  is available at \url{https://github.com/iWeisskohl.}}}. The experimental settings are first described in Section \ref{subsec:settings}. A sensitivity analysis with respect to the main hyperparameters is first reported in Section \ref{subsec:sensitivity}. We then compare the segmentation  accuracy and calibration of our models with those of  state-of-the-art models in Sections  \ref{subsec:state-of-art} and \ref{subsec:calibration}, respectively.

\subsection{Experimental settings}
\label{subsec:settings}
\paragraph{Dataset}
The dataset considered in this paper contains 3D images from 173 patients who were diagnosed with large B-cell lymphomas and underwent PET-CT examination. (The study was approved as a retrospective study by the Henri Becquerel Center Institutional Review Board). The lymphomas in mask images were delineated manually by experts and considered as ground truth. All PET/CT data were stored in the  DICOM (Digital Imaging and Communication in Medicine) format. The size and spatial resolution of PET and CT images and the corresponding mask images vary due to the use of different imaging machines and operations. For CT images, the size varies from $267\times 512\times512$ to $478\times 512\times512$. For PET images, the size varies from $276\times 144\times144$ to $407\times 256\times256$. 

\paragraph{Pre-processing}
Several pre-processing methods were used to process the PET/CT data. At first, the data in DICOM format were transferred into the NIFTI (Neuroimaging Informatics Technology Initiative) format for further processing. Second, the PET, CT and mask images were normalized: (1) for PET images, we applied a random intensity shift and scale of each channel with the shift value of 0 and scale value of 0.1; (2) for CT images, the shift and scale values were set to 1000 and 1/2000; (3) for mask images, the intensity value was normalized into the $[0,1]$ interval by replacing the outside value by $1$. Third, PET and CT images were resized to $256\times 256\times 128$ by linear interpolation, and mask images were resized to $256\times 256\times 128$ by nearest neighbor interpolation. Lastly,
the registration of CT and PET images was performed by B-spline interpolation. All the prepossessing methods can be found in the SimpleITK \cite{lowekamp2013}\cite{yaniv2018} toolkit. During training, PET and CT images were concatenated as a two-channel input. We randomly selected 80\% of the data for training, 10\% for validation and 10\% for testing. \new{This partition was fixed and used in all the experiments reported below.}  

\paragraph{Parameter initialization}
For the evidential layer module, we considered two variants based on the ENN classifier recalled in Section \ref{subsec:enn} on the one hand, and on an RBF network as described in Section \ref{subsec:nnweights} \new{on the other hand}. Both approaches are based on prototypes in the space of features extracted by the UNet module. When using ENN or RBF classifiers as stand-alone classifiers, prototypes are usually initialized by a clustering algorithm such as the $k$-means. Here, this approach is not so easy, because the whole network is trained in an end-to-end way, and the features are constructed during the training process. However, $k$-means initialization can still be performed by a four-step process:
\begin{enumerate}
    \item A standard UNet architecture (with a softmax output layer) is trained end-to-end;
    \item The $k$-means algorithm is run in the space of features extracted by the trained UNet;
    \item The evidential layer is trained alone, starting from the initial prototypes computed by the $k$-means;
    \item The whole model (feature extraction module and evidential layer) is fine-tuned by end-to-end learning with a small learning step.
\end{enumerate}
As an alternative method, we also considered \new{training the feature extraction module and the evidential layer  simultaneously, in which case the prototypes were initialized randomly} from a normal distribution with zero mean and identity covariance matrix. For the ENN module, the initial values of parameters $\alpha_i$ and $\gamma_i$ were set, respectively, at 0.5 and 0.01, and membership degrees $u_{ik}$ were initialized randomly by drawing uniform random numbers and normalizing. For the RBF module, the initial value of the scale parameter \textsf{$\gamma_i$} of RBF was set to 0.01, and the weight $v_i$ were drawn randomly from a standard normal distribution.

\paragraph{Learning algorithm} Each model was trained on the learning set with 100 epochs using the Adam optimization algorithm. The initial learning rate was set to $10^{-3}$. An adjusted learning rate schedule was applied by reducing the learning rate when the training loss did not decrease in 10 epochs. The model with the best performance on the validation set was saved as the final model for testing.  
All methods were implemented in Python with the PyTorch-based medical image framework MONAI, and were trained and tested on a desktop with a 2.20GHz Intel(R) Xeon(R) CPU E5-2698 v4 and a Tesla V100-SXM2 graphics card with 32 GB GPU memory. 

\begin{figure}
\centering
\includegraphics[width=0.4\textwidth]{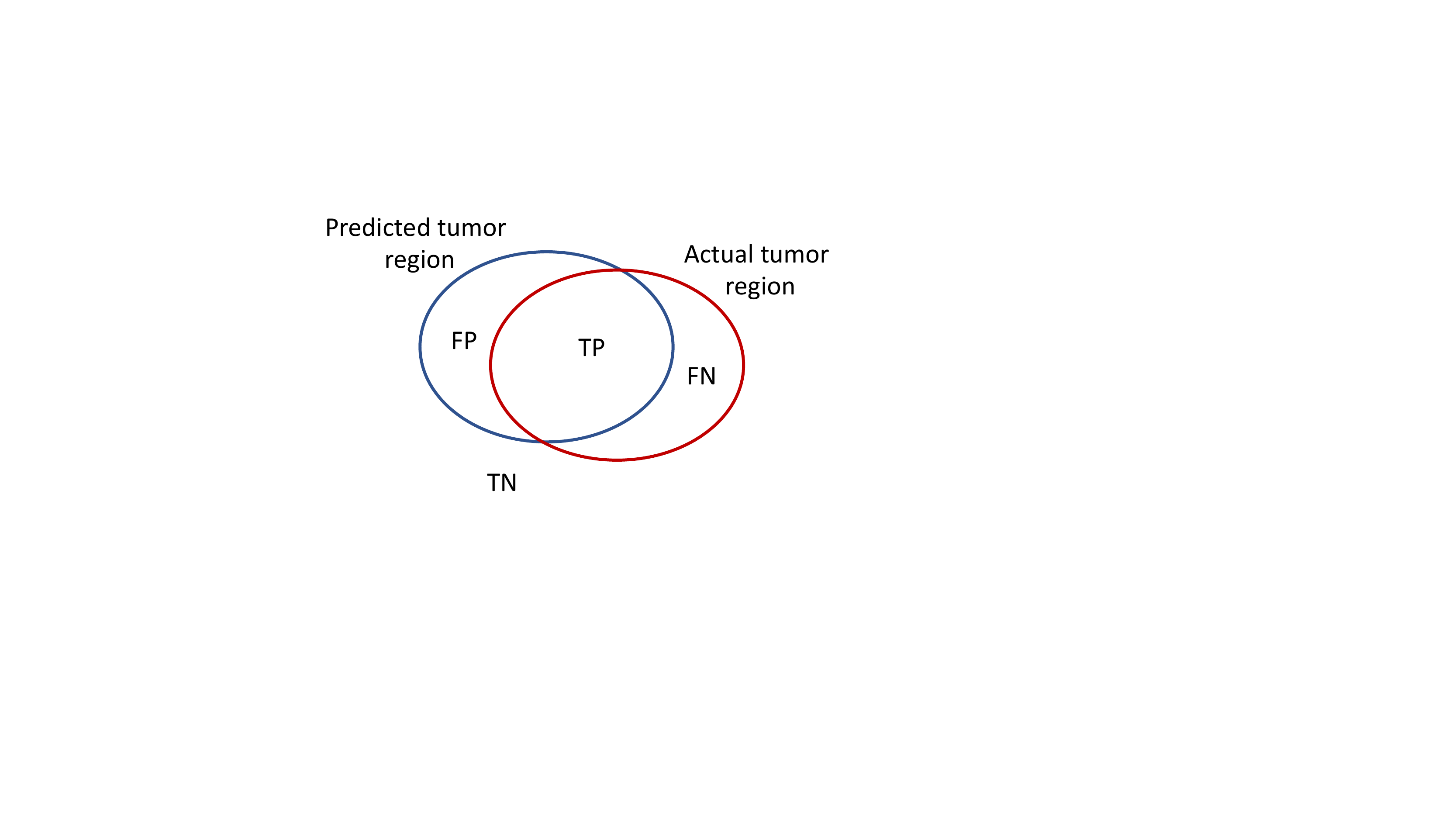}
\caption{Geometric interpretation of the numbers of true positive (TP), false positive (FP), true negative (TN) and false negative (TN) used for the definition of evaluation criteria. \label{fig:criteria}}
\end{figure}

\paragraph{Evaluation criteria}
The evaluation criteria most commonly used to assess the quality of medical image segmentation algorithms are the Dice score, Sensitivity and Precision. These criteria are defined as follows:
\begin{equation*}
    \textsf{Dice}(P,T)=\frac{2\times TP}{FP+2\times TP+FN}, 
\end{equation*}
\begin{equation*}
    \textsf{Sensitivity}(P,T)=\frac{TP}{TP+FN},
\end{equation*}
\begin{equation*}
    \textsf{Precision}(P,T)=\frac{TP}{TP+FP},
\end{equation*}
where $TP$, $FP$, and $FN$ denote, respectively, the numbers of true positive, false positive, false negative voxels (See Figure \ref{fig:criteria}). The reported results in the following sections were obtained by calculating these three criteria for each test 3D image and then averaging over the patients. The Dice score is a global measure of  segmentation performance. It is equal to twice the volume of the intersection between the predicted and actual tumor regions, divided by the sum of the volumes of these regions. 
Sensitivity is the proportion, among actual tumor voxels, of voxels  correctly predicted as tumor. Precision is the proportion, among predicted tumor voxels, of voxels that actually belong to the tumor region; it is, thus, an estimate of the probability that the model is correct when it predicts that a voxel is in a lymphoma region. We note that neither sensitivity, nor precision are global performance criteria. We can increase sensitivity by predicting the tumor class more often (at the expense of misclassifying a lot of background pixels), and we can increase precision by being very cautious and predicting the tumor class only when it has a high probability (at the expense of missing a lot of tumor voxels). These two criteria, thus, have to be considered jointly. Finally, we can also remark that a forth criterion can also be defined: specificity, which is the proportion, among background voxels, of voxels correctly predicted as background (i.e., $TN/(TN+FP)$). However, as there are much more background voxels than tumor ones, this criterion is not informative in tumor segmentation applications (it is always very close to 1). 

\new{In addition to quality of the  segmentation,  we also wish to evaluate the calibration of output probabilities or belief functions (see Section \ref{subsec:calibration}). For that purpose, we will use an additional evaluation criterion, the Expected Calibration Error (ECE) \cite{guo2017calibration}. The output pignistic probabilities from the evidential layer are first discretized into $R$ equally spaced bins $B_r$, $r=1,\ldots,R$ (we used $R=10$). The accuracy of bin $B_r$ is defined as
 \begin{equation}
     \acc(B_r)=\frac{1}{\mid B_r \mid}\sum_{i\in B_r}^{} \boldsymbol{1} (P_i=G_i),
 \end{equation}
 where $P_i$ and $G_i$ are, respectively, the predicted and true class labels for sample $i$. The average confidence of bin $B_r$ is defined as
 \begin{equation}
\conf(B_r)=\frac{1}{\mid B_r \mid}\sum_{i\in B_r}^{}S_i,
 \end{equation}
where $S_i$ is the confidence for sample $i$. The ECE is the weighted average of the difference in accuracy and confidence of the bins:
\begin{equation}
\label{eq:ECE}
\textsf{ECE}= \sum_{r=1}^{R} \frac{\mid B_r \mid }{N}\mid \acc(B_r)-\conf(B_r)\mid,
\end{equation}
where $N$ is the total number of elements in all bins, 
and $\mid B_r\mid$ is the number of elements in bin $B_r$. A model is perfectly calibrated when $\acc(B_r)=\conf(B_r)$ for all $r\in \{1,\ldots,R\}$. Through the bin-size weighting in the ECE metric, the highly confident and accurate background voxels significantly affect the results. Because  our dataset has imbalanced foreground and background proportions, we  only considered voxels belonging to the tumor to calculate the ECE, similar to \cite{jungo2020analyzing}\cite{rousseau2021post}. For each patient in the test set, we defined a bounding box covering the lymphoma region and calculated the ECE in this bounding box. We are interested in the patient-level ECE and thus reported the mean patient ECE instead of the voxel-level ECE (i.e., considering all voxels in the test set to calculate the ECE).
}

\subsection{Sensitivity analysis}
\label{subsec:sensitivity}

We analyzed the sensitivity of the results to the main design hyperparameters, which are: the number $H$ of extracted features, the number $I$ of prototypes and the regulation coefficient $\lambda$. The influence of the initialization method was also be studied. \new{In all the experiments reported in this section as well as in Section \ref{subsec:state-of-art}, learning in each of the configurations was repeated five times with different random initial conditions.}

\paragraph{Influence of the number of features}
Table \ref{tab:input_dim} shows the means and standard deviations (over five runs) of the three performance indices for ENN-UNet and RBF-UNet with different numbers of features ($H\in\{2,5,8\}$). The number of prototypes and the regularization coefficient were set, respectively, to $I=10$ and $\lambda=0$. The prototypes were initialized randomly. ENN-UNet achieves the highest Dice score and sensitivity with $H=2$ features, but the highest precision with $H=8$. However, the differences are small and concern only the third decimal point. Similarly, RBF-UNet had the best values of the Dice score and precision for $H=5$ features, but again the differences are small. Overall, it seems that only two features are sufficient to discriminate between tumor and background voxels.

\begin{table}
\caption{Means and standard deviations (over five runs) of the performance measures for different input dimensions $H$, with $I=10$ randomly initialized prototypes and  $\lambda=0$. The best values are shown in bold.}
\centering
\label{tab:input_dim}
\begin{tabular}{cccccccc}
\hline
 Model &$H$ &\multicolumn{2}{c}{Dice score} &\multicolumn{2}{c}{Sensitivity}&  \multicolumn{2}{c}{Precision} \\
\cline{3-8} 
&&Mean&SD&Mean&SD&Mean&SD\\
\hline
\multirow{3}*{ENN-UNet} & 2  & \textbf{0.833}&0.009 & \textbf{0.819} &0.019& 0.872&0.018
\\
&5& 0.831 &0.012 &0.817&0.016 &0.870&0.011\\
&8  &0.829& 0.006
&0.816& 0.010 & \textbf{0.877}& 0.019\\
\hline
\multirow{3}*{RBF-UNet}&2& 0.824&0.009&\textbf{0.832}&0.008&  0.845 &0.016\\
&5&  \textbf{0.825}&0.006&0.817&0.016&\textbf{0.862}&0.010\\
&8& 0.821&0.011&0.813&0.010&  0.862&0.022\\
\hline
\end{tabular}
\end{table}

\paragraph{Influence of the regularization coefficient}
In the previous experiment, the networks were trained without regularization. Tables \ref{tab:lambda1} and \ref{tab:lambda2} show the performances of ENN-UNet and RBF-UNet for different values of $\lambda$, with $I=10$ randomly initialized prototypes and, respectively, $H=2$ and $H=8$ inputs. With both settings, ENN-UNet does not benefit from regularization (the best results are obtained with $\lambda=0$). In contrast, RBF-UNet is more sensitive to regularization, and achieves the highest Dice score  with  $\lambda=0.01$. This finding confirms the remark already made in Section \ref{subsec:compar}, where it was observed that an ENN classifier seems to be less sensitive to regularization than an RBF classifier (see Figure \ref{fig:error}).

\begin{table}
\caption{Means and standard deviations (over five runs) of the performance measures for  different values of the regularization coefficient $\lambda$, with $I=10$ randomly initialized prototypes and $H=2$ features. The best values are shown in bold.}
\centering
\label{tab:lambda1}
\begin{tabular}{ccccccccc}
\hline
 Model &$\lambda$ &\multicolumn{2}{c}{Dice score} &\multicolumn{2}{c}{Sensitivity}&  \multicolumn{2}{c}{Precision} \\
\cline{3-8} 
 & &Mean&SD&Mean&SD&Mean&SD\\
\hline
\multirow{3}*{ENN-UNet} & 0  &\textbf{0.833} &0.009&\textbf{0.819} & 0.019&\textbf{0.872}&0.018\\
&1e-4& 0.822&	0.007&0.818&0.026&	0.839&0.035\\
&1e-2  &0.823&0.004&0.817&0.023	&0.856&0.023\\
\hline
\multirow{3}*{RBF-UNet}&0&0.824&0.009& \textbf{0.832}&0.008&0.845&0.016\\
&1e-4& 0.825&	0.011&0.811&0.022&	\textbf{0.869}&0.020\\
&1e-2& \textbf{0.829}&0.010&0.818	&0.022 &0.867&0.016\\
\hline
\end{tabular}
\end{table}

\begin{table}
\caption{Means and standard deviations (over five runs) of the performance measures for  different values of the regularization coefficient $\lambda$, with $I=10$ randomly initialized prototypes and  $H=8$ features. The best values are shown in bold.}
\centering
\label{tab:lambda2}
\begin{tabular}{ccccccccc}
\hline
 Model &$\lambda$ &\multicolumn{2}{c}{Dice score} &\multicolumn{2}{c}{Sensitivity}&  \multicolumn{2}{c}{Precision} \\
\cline{3-8} 
 & &Mean&SD&Mean&SD&Mean&SD\\
\hline
\multirow{3}*{ENN-UNet} & 0  &\textbf{0.829} &0.006&\textbf{0.811} & 0.010&\textbf{0.877}&0.019\\
&1e-4& 0.827&	0.008&0.809&0.019&0.873&0.024\\
&1e-2  &0.822&0.009&	0.807&0.021	&0.867&0.011\\
\hline
\multirow{3}*{RBF-UNet}&0&0.821&0.010&0.813&0.010&0.862&0.022\\
&1e-4& 0.827&	0.004&\textbf{0.830}& 0.005&	0.852& 0.012\\
&1e-2& \textbf{0.832}&0.006&0.825	&0.022&  \textbf{0.867}&0.020\\
\hline
\end{tabular}
\end{table}

\paragraph{Influence of the number of prototypes}
The number $I$ of prototypes is another hyperparameter that may impact segmentation performance. Table \ref{tab:protos} shows the performances of ENN-UNet and RBF-UNet with 10 and 20 randomly initialized prototypes, the other hyperparameters being fixed at  $H=2$ and $\lambda=0$. Increasing the number of prototypes beyond 10 does not seem to improve the performance of ENN-UNet, while it does slightly improve the performance of RBF-UNet in terms of Dice score and precision, at the expense of an increased computing time.

\begin{table}
\caption{Means and standard deviations (over five runs) of the performance measures for  different numbers $I$ of randomly initialized prototypes, with $H=2$ features and  $\lambda=0$. The best values are shown in bold.}
\centering
\label{tab:protos}
\begin{tabular}{cccccccccc}
\hline
 Model &$I$ &\multicolumn{2}{c}{Dice score} &\multicolumn{2}{c}{Sensitivity}&  \multicolumn{2}{c}{Precision} \\
\cline{3-8} 
 &&Mean&SD&Mean&SD&Mean&SD\\
\hline
\multirow{2}*{ENN-UNet} & $10$  & \textbf{0.833} &0.009&\textbf{0.819} & 0.019&\textbf{0.872}&0.018\\
&$20$& 0.823&0.007&0.804&0.006&0.864&0.012\\
\hline
\multirow{2}*{RBF-UNet}&$10$&0.824&0.009& \textbf{0.832}&0.008&0.845&0.016\\
&$20$& \textbf{0.830}&0.007& 0.810&0.012&\textbf{0.867}&0.010\\
\hline
\end{tabular}
\end{table}

\paragraph{Influence of the prototype initialization method}

Finally, we compared the two initialization methods mentioned in Section \ref{subsec:settings}. For $k$-means initialization, in the first step, a UNet model was trained with the following settings: kernel size=$5$, channels =$(8, 16, 32, 64, 128)$ and strides=$(2,2,2,2)$. The spatial dimension, input and output channel were set, respectively, 3, 2, and 2. This  pre-trained UNet was used to extract $H=2$ features, and 10 prototypes were obtained by running the $k$-means algorithm in the space of  extracted features. These prototypes were fed into ENN or RBF layers, which were trained separately, with fixed features. For this step, the  learning rate was set to $10^{-2}$. Finally, the whole model was fine-tuned end-to-end, with  a smaller learning rate equal to $10^{-4}$. 
Table \ref{tab:initial} shows the performances of ENN-UNet and RBF-UNet with random and $k$-means initialization. Both ENN-UNet and RBF-UNet achieve a higher Dice score when using the $k$-means initialization method, and the variability of the results is also reduced with this method.

\begin{table}
\caption{Means and standard deviations (over five runs) of the performance measures for different initialization methods, with $I=10$ prototypes, $H=2$ features and $\lambda=0$. The best values are shown in bold.}
\centering
\label{tab:initial}
\begin{tabular}{cccccccccc}
\hline
 Model & Initialization&\multicolumn{2}{c}{Dice score} &\multicolumn{2}{c}{Sensitivity}&  \multicolumn{2}{c}{Precision} \\
\cline{3-8} 
 & &Mean&SD&Mean&SD&Mean&SD\\
\hline

\multirow{2}*{ENN-UNet} &Random &0.833 &0.009&0.819 & 0.019&0.872&0.018\\
&  $k$-means&\textbf{0.846}&0.002&\textbf{0.830}&0.004&\textbf{0.879}&0.008
\\

\hline
\multirow{2}*{RBF-UNet}&Random&0.824&0.009& \textbf{0.832}&0.008&0.845&0.016\\
&  $k$-means&\textbf{0.839}&0.003&0.824&0.001&\textbf{0.879}&0.008\\

\hline
\end{tabular}
\end{table}

\begin{figure}
\centering
\subfloat[\label{fig:contourESomega1}]{\includegraphics[width=0.5\textwidth]{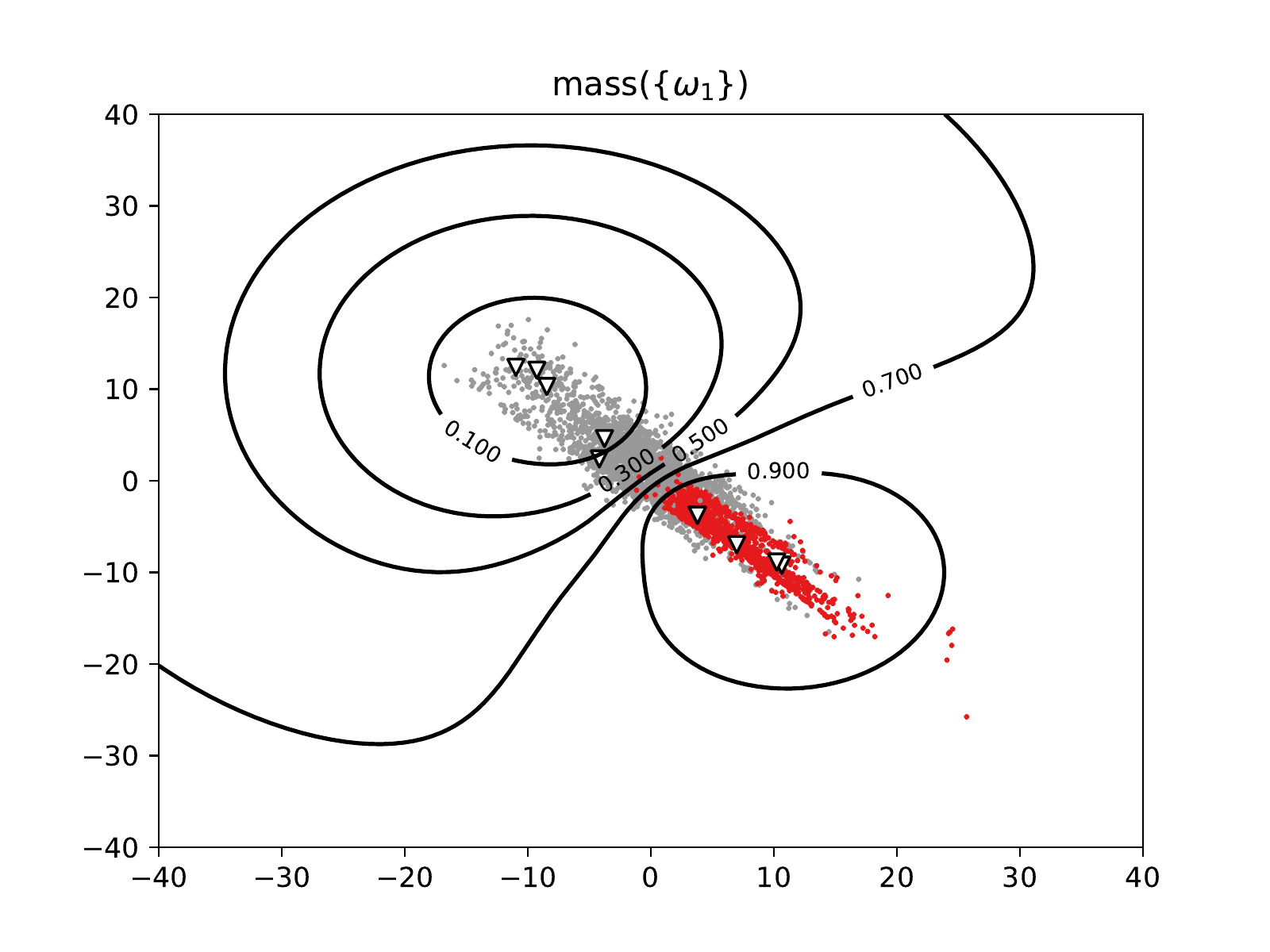}}
\subfloat[\label{fig:contourESomega2}]{\includegraphics[width=0.5\textwidth]{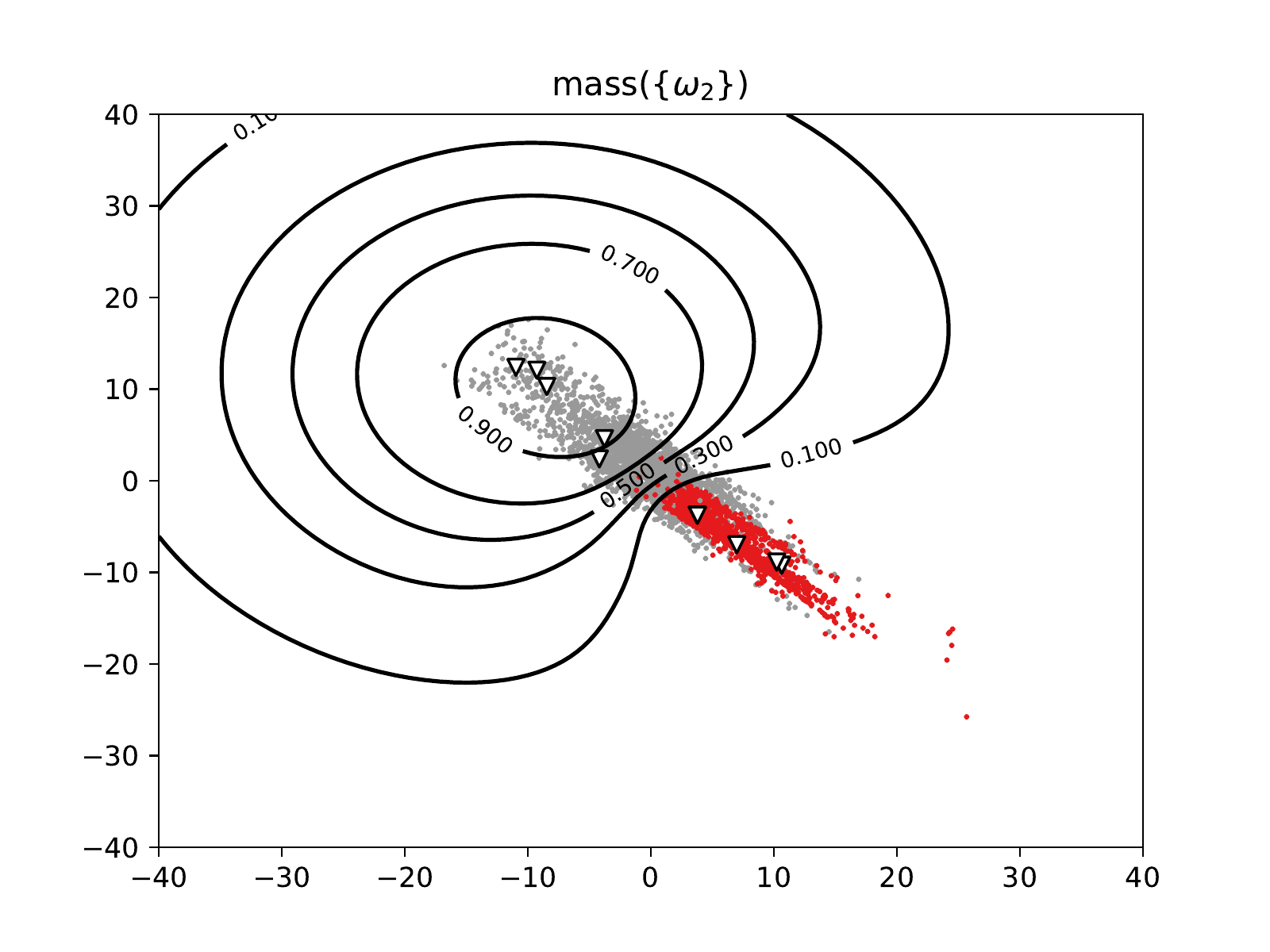}}\\
\subfloat[\label{fig:contourESOmega}]{\includegraphics[width=0.5\textwidth]{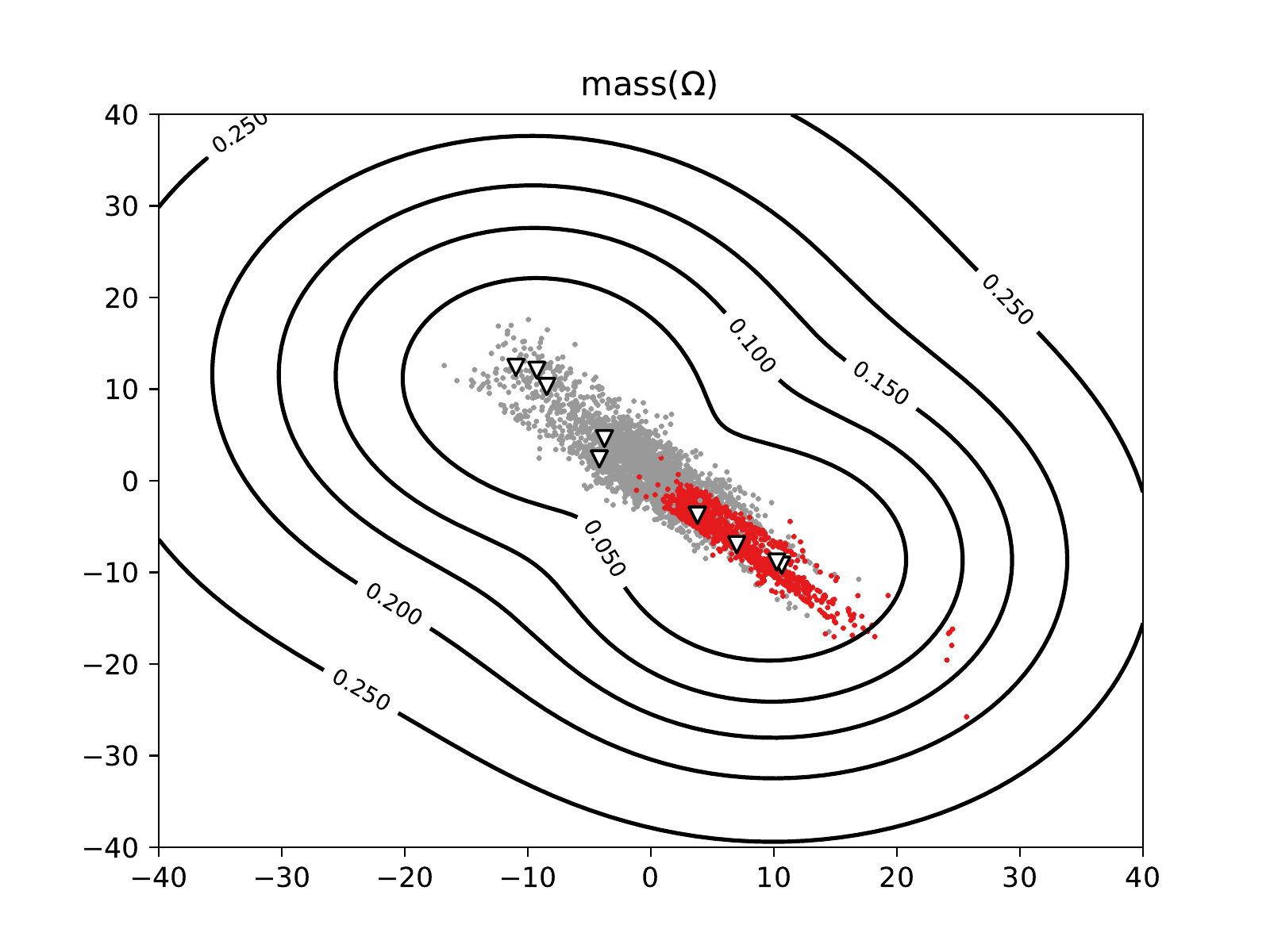}}
\caption{Contours in feature space of the masses assigned to the background (a), the tumor class (b) and the frame of discernment (c) by the ENN-UNet model initialized by $k$-means. Training was done with $\lambda=10^{-2}$, $H=2$ and $I=10$. Sampled feature vectors from the tumor and background classes are marked in gray and red, respectively. 
\label{fig:ES-kmeans}}
\end{figure}

\begin{figure}
\centering
\subfloat[\label{fig:contourRBFUomega1}]{\includegraphics[width=0.5\textwidth]{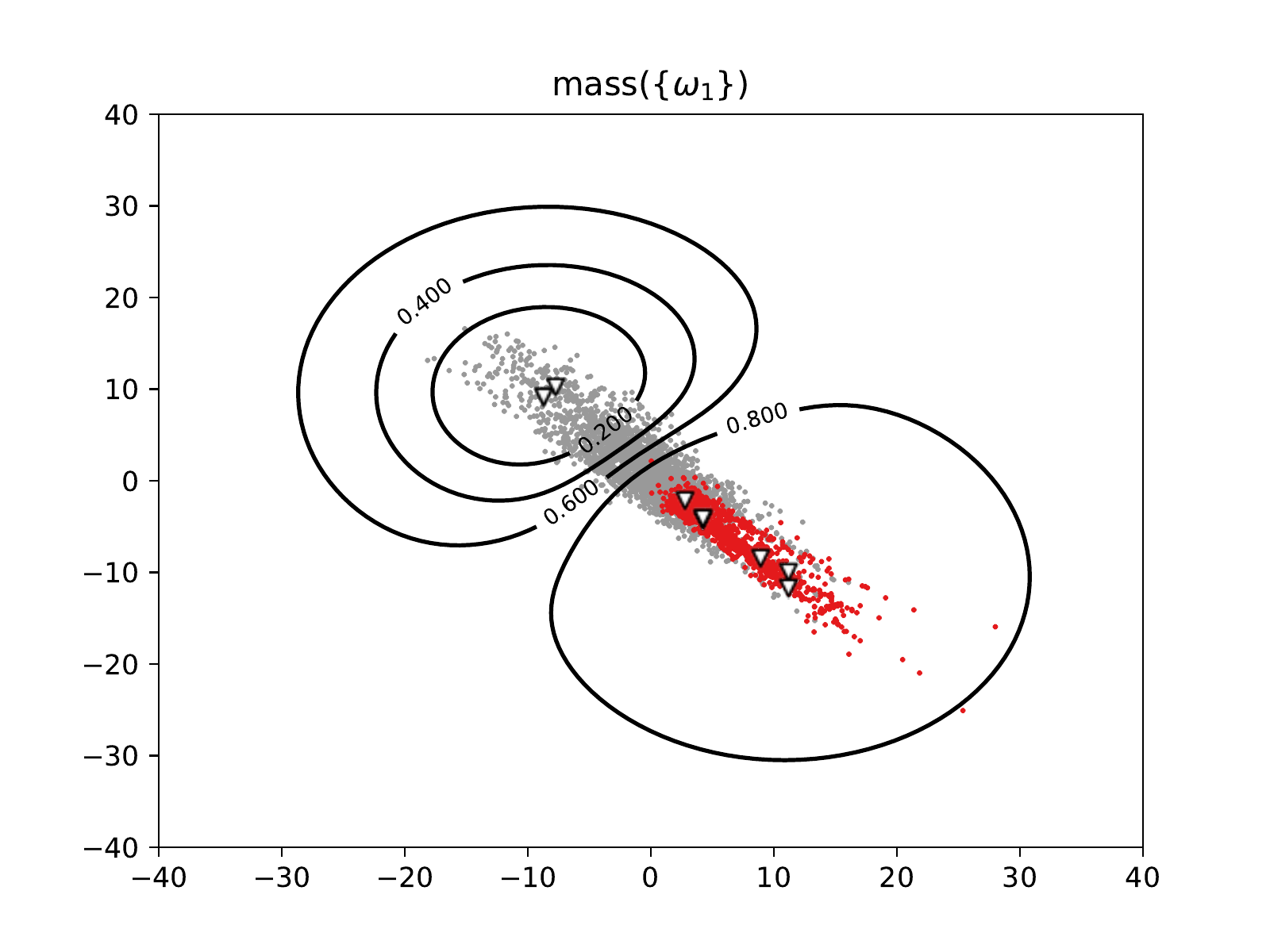}}
\subfloat[\label{fig:contourRBFUomega2}]{\includegraphics[width=0.5\textwidth]{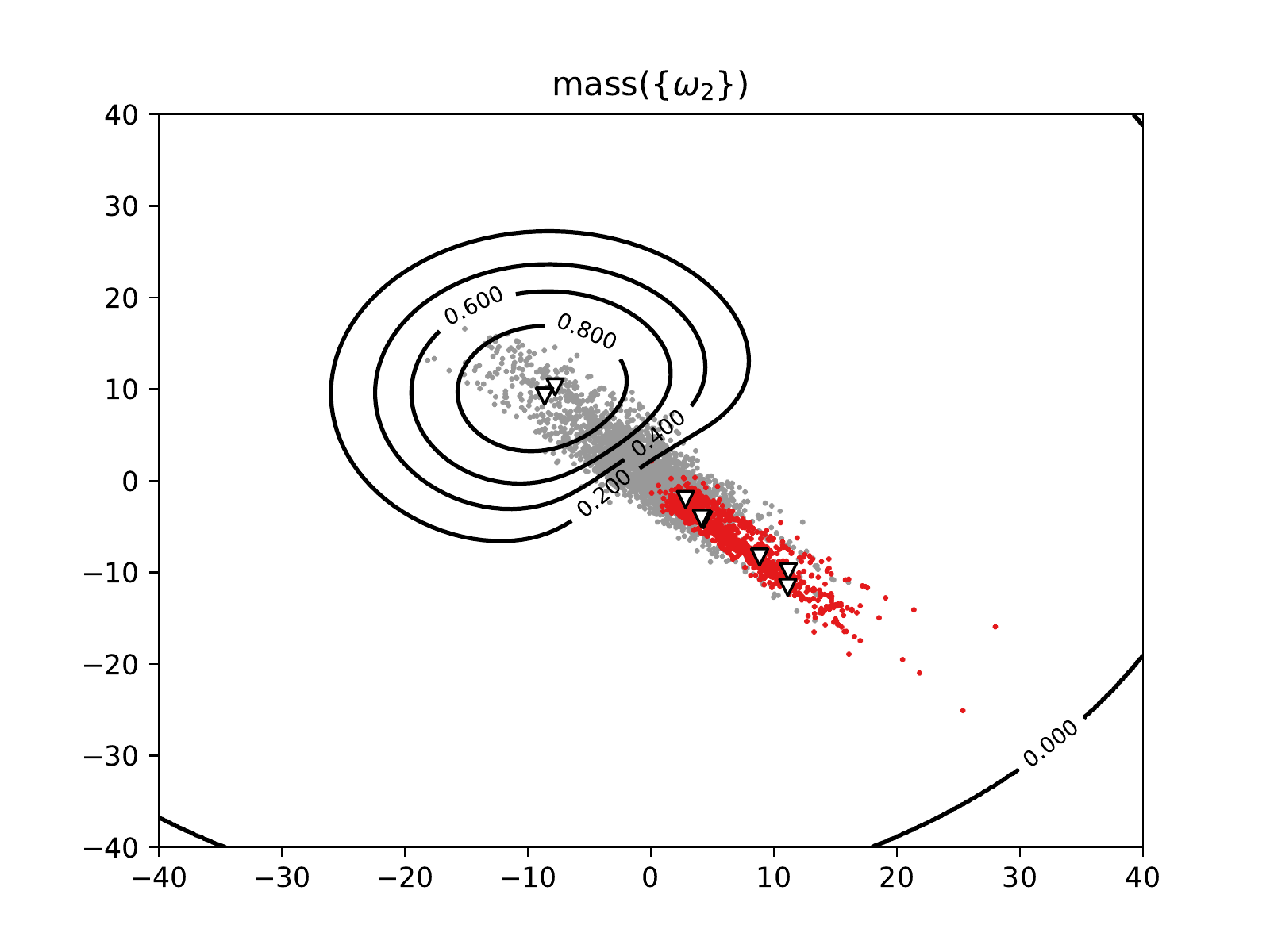}}\\
\subfloat[\label{fig:contourRBFUOmega}]{\includegraphics[width=0.5\textwidth]{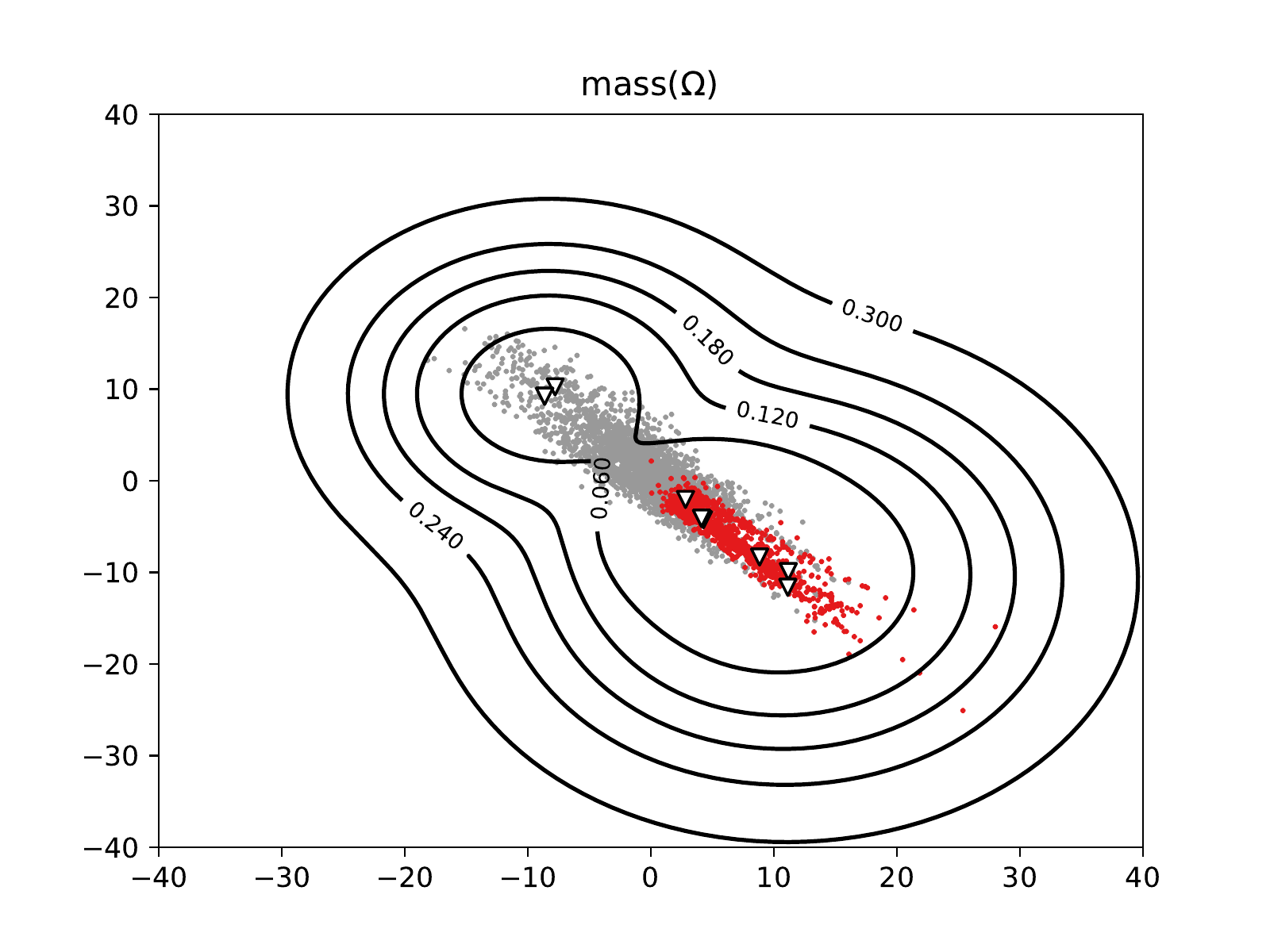}}
\caption{Contours in feature space of the masses assigned to the background (a), the tumor class (b) and the frame of discernment (c) by the RBF-UNet model initialized by $k$-means. Training was done with $\lambda=10^{-2}$, $H=2$ and $I=10$. Sampled feature vectors from the tumor and background classes are marked in gray and red, respectively. 
\label{fig:RBF-kmeans}}
\end{figure}

\begin{figure}
\centering
\subfloat[\label{fig:contourESRomega1}]{\includegraphics[width=0.5\textwidth]{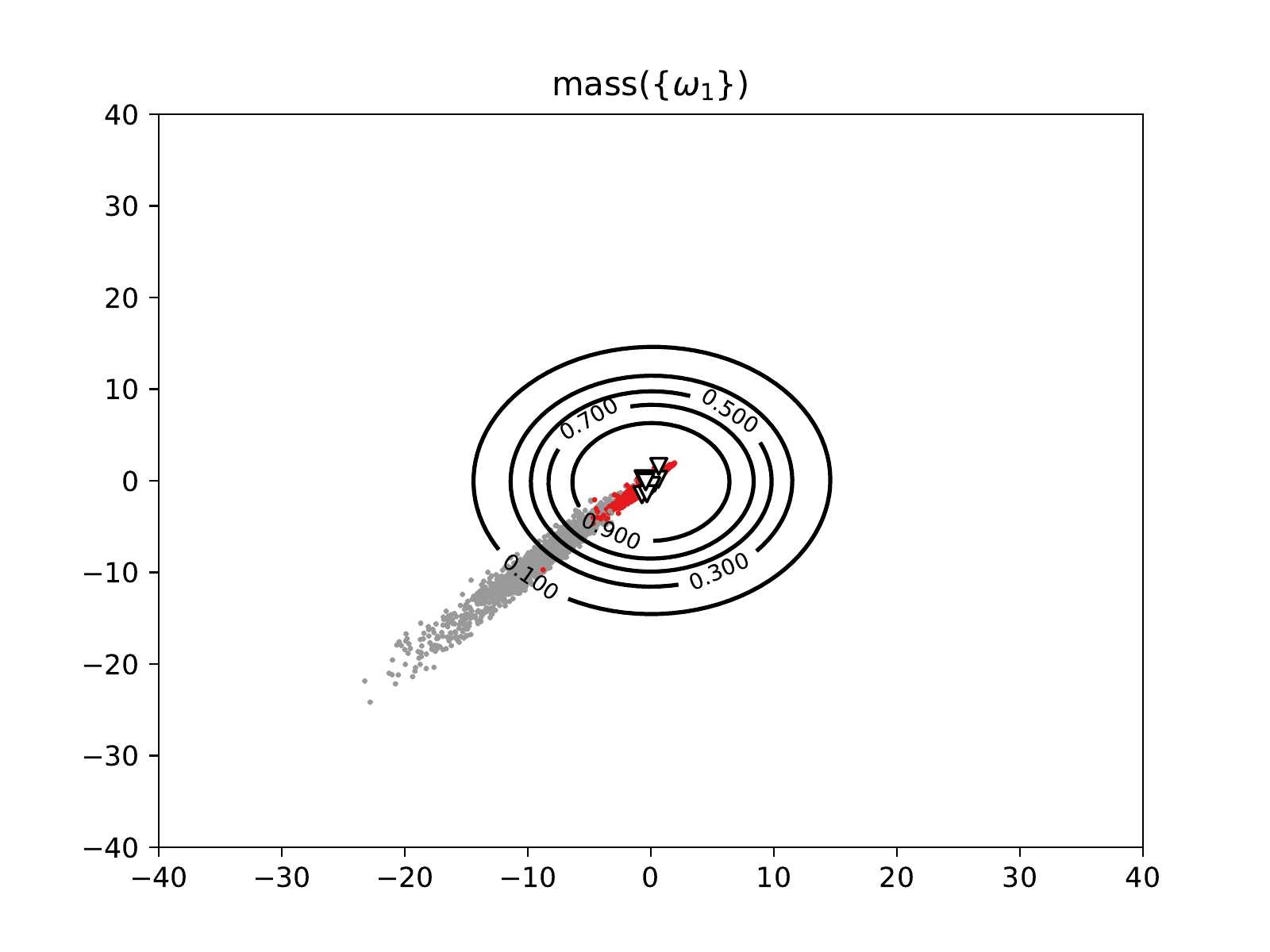}}
\subfloat[\label{fig:contourESRomega2}]{\includegraphics[width=0.5\textwidth]{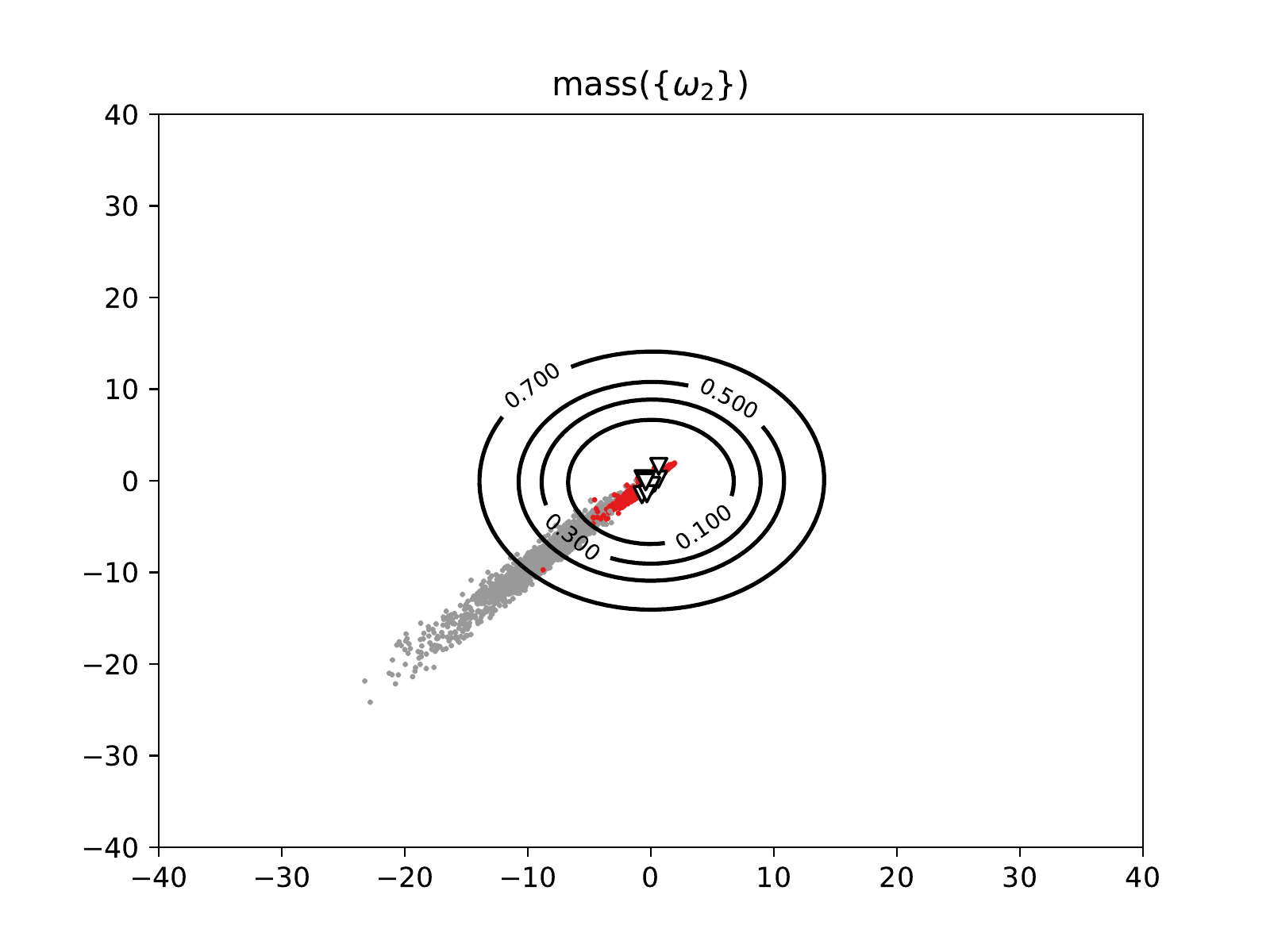}}\\
\subfloat[\label{fig:contourESROmega}]{\includegraphics[width=0.5\textwidth]{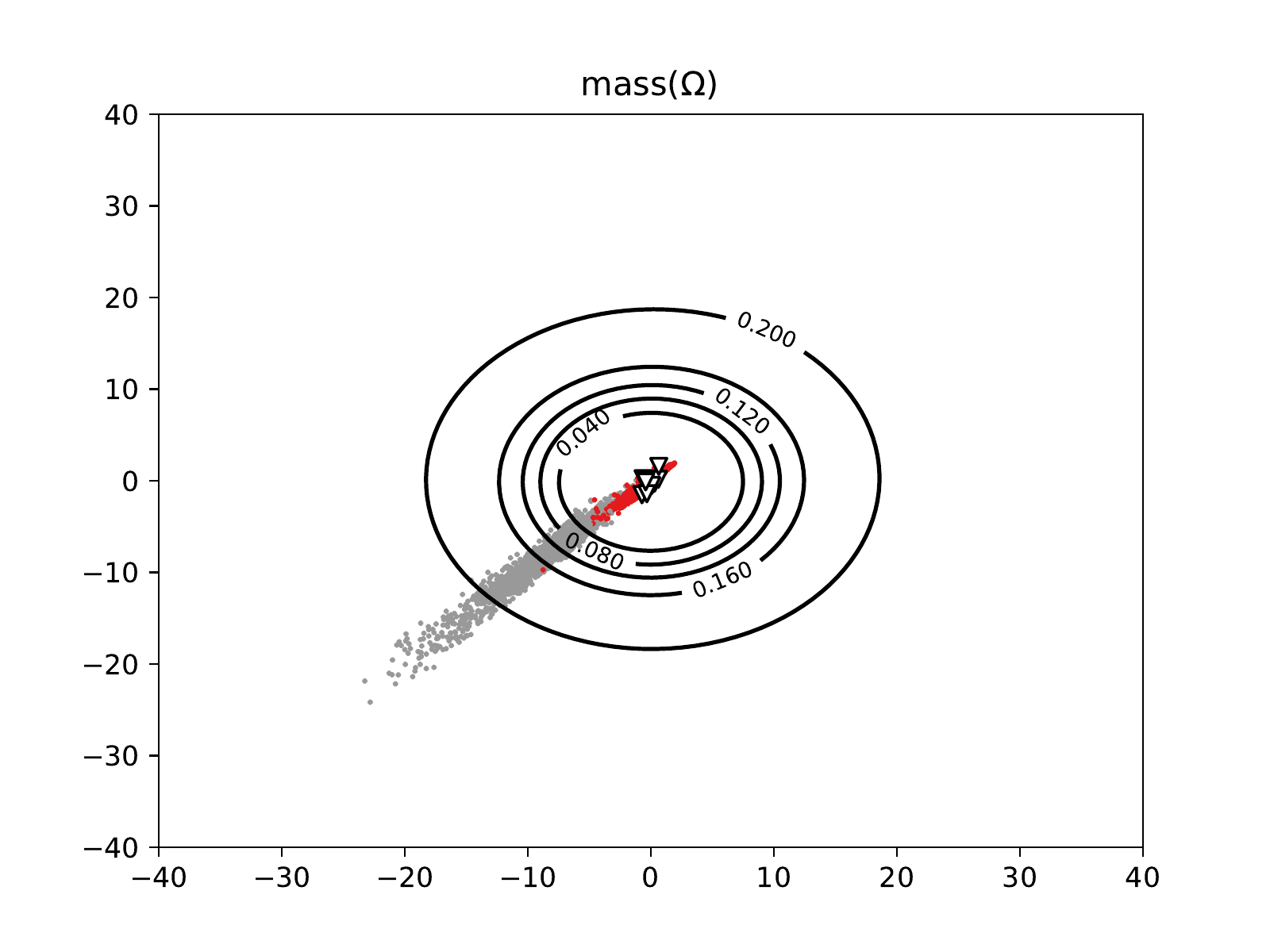}}
\caption{Contours in feature space of the masses assigned to the background (a), the tumor class (b) and the frame of discernment (c) by the ENN-UNet model initialized randomly. Training was done with $\lambda=10^{-2}$, $H=2$ and $I=10$. Sampled feature vectors from the tumor and background classes are marked in gray and red, respectively. 
\label{fig:ES-random}}
\end{figure}

Not only does the $k$-means initialization method slightly improve the performances of ENN-UNet and RBF-UNet quantitatively, but it also \new{tends to position the prototypes in regions of high data density. As a result, a high  output mass $m(\Omega)$ signals that the input data is atypical. In that sense, the output mass function is more interpretable}. This point is illustrated by Figures \ref{fig:ES-kmeans} and \ref{fig:RBF-kmeans}, which show the contours, in the two-dimensional feature space, of the masses assigned to the background, the tumor class and the frame of discernment when using $k$-means initialization (with $\lambda=10^{-2}$ and $I=10$) with, respectively, ENN-UNet and RBF-Unet.  For both models, the prototypes are well distributed over the two classes, and the mass on $\Omega$ decreases with the distance to the data, as expected. In contrast, when using random initialization (as shown in Figure \ref{fig:ES-random} for the ENN-UNet model -- results are similar with the RBF-UNet model), the prototypes  are located in the background region, \new{and the mass $m(\Omega)$ does not have a clear meaning} (although the decision boundary still ensures a good discrimination between the two classes).

From this sensitivity analysis, we can conclude that the performances of both ENN-UNet and RBF-UNet are quite robust to the values of the hyperparameters, and that the two models achieve comparable performances. The $k$-means initialization method seems to yield better results, both quantitatively and qualitatively. The next section is devoted to a comparison with alternative models.


\subsection{Comparative analysis: segmentation accuracy} 
\label{subsec:state-of-art}

In this section, we compare the performances of the ENN-UNet and RBF-UNet models with those of the baseline model, UNet \cite{ronnebergerconvolutional}, as well as three state-of-the-art models reviewed in Section \ref{intro}: VNet \cite{milletari2016v}, SegResNet \cite{myronenko20183d} and nnUNet \cite{isensee2018nnu}.  For all compared methods, the same learning set and pre-processing steps were used. \new{All the compared methods were trained with the Dice loss function \eqref{eq:Dice_loss}.} Details about the optimization algorithm were given in Section \ref{subsec:settings}. All methods were implemented based on the MONAI framework\footnote{More details about how to use those models can be found from MONAI core tutorials \url{https://monai.io/started.html##monaicore}.} and can be called directly. For UNet, the kernel size was set as 5 and the channels were set to $(8, 16, 32, 64, 128)$ with strides=$(2,2,2,2)$. 
For nnUNet, the kernel size was set as $(3, (1,1, 3), 3, 3)$ and the upsample kernel size was set as $(2,2,1)$ with strides $((1,1,1), 2, 2, 1)$. For SegResNet \cite{myronenko20183d} and VNet \cite{milletari2016v}, we used the pre-defined model without changing any parameter. The spatial dimension, input channel and output channel were set, respectively, 3, 2, and 2 for the four compared models. \new{
As for other hyperparameters not mentioned here, we used the pre-defined value given in MONAI.} As shown by the sensitivity analysis performed in Section \ref{subsec:sensitivity}, the best results for ENN-UNet and RBF-UNet are achieved with  $\lambda=0$, $I=10$, $H=2$ and $k$-means initialization.

\begin{figure}
\centering
\subfloat[\label{fig:dotplot_Dice}]{\includegraphics[width=0.4\textwidth]{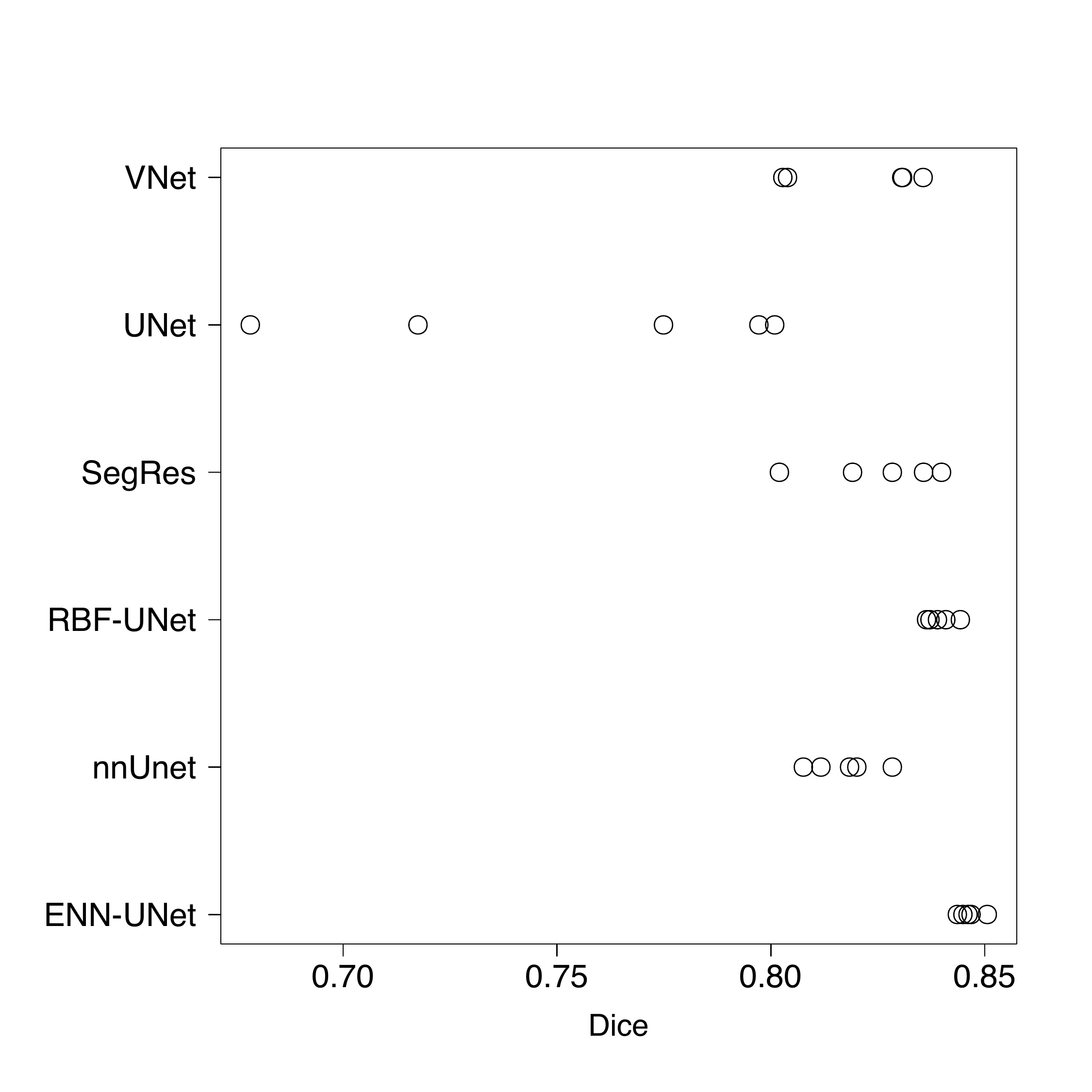}}
\subfloat[\label{fig:dotplot_Sensitivity}]{\includegraphics[width=0.4\textwidth]{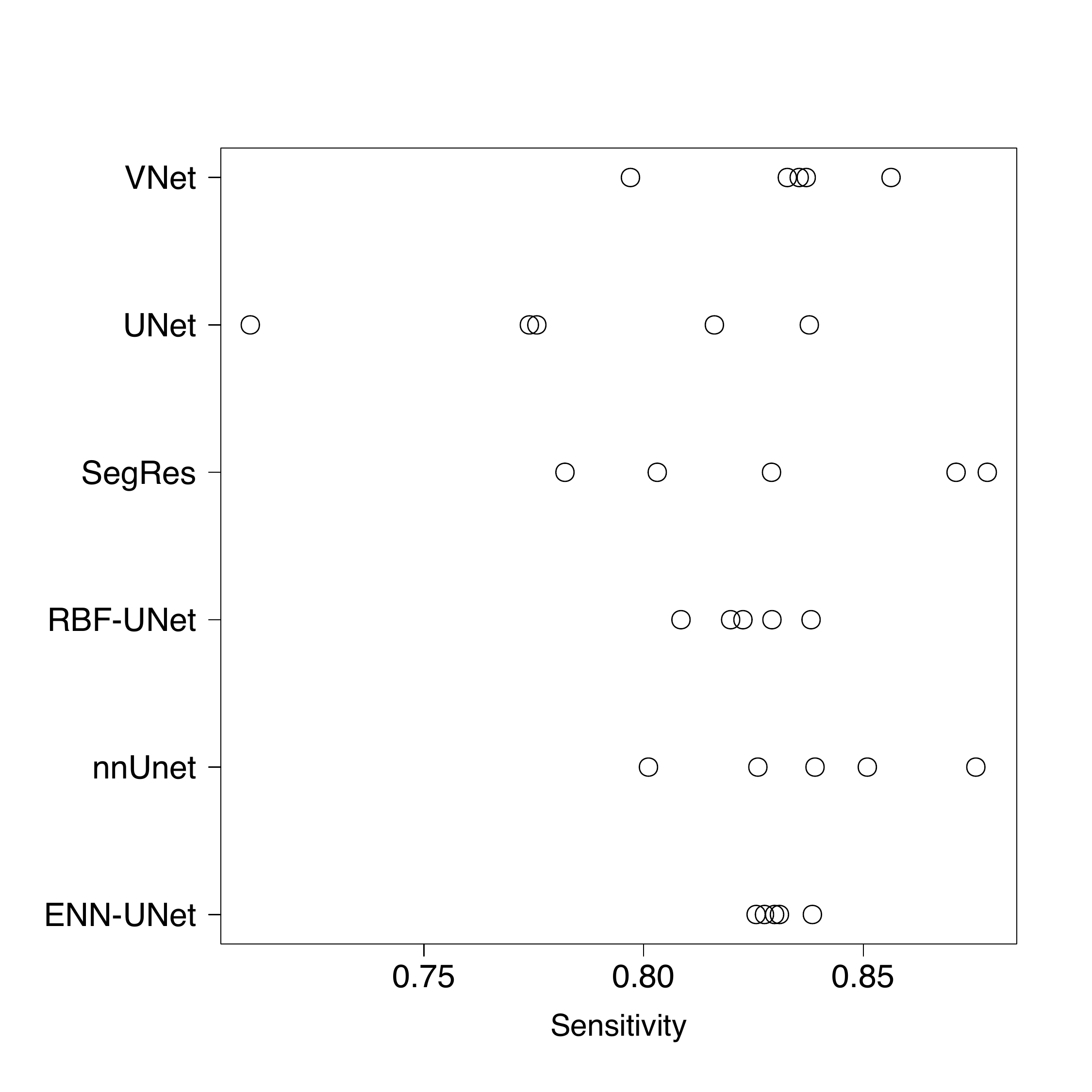}}\\
\subfloat[\label{fig:dotplot_Precision}]{\includegraphics[width=0.4\textwidth]{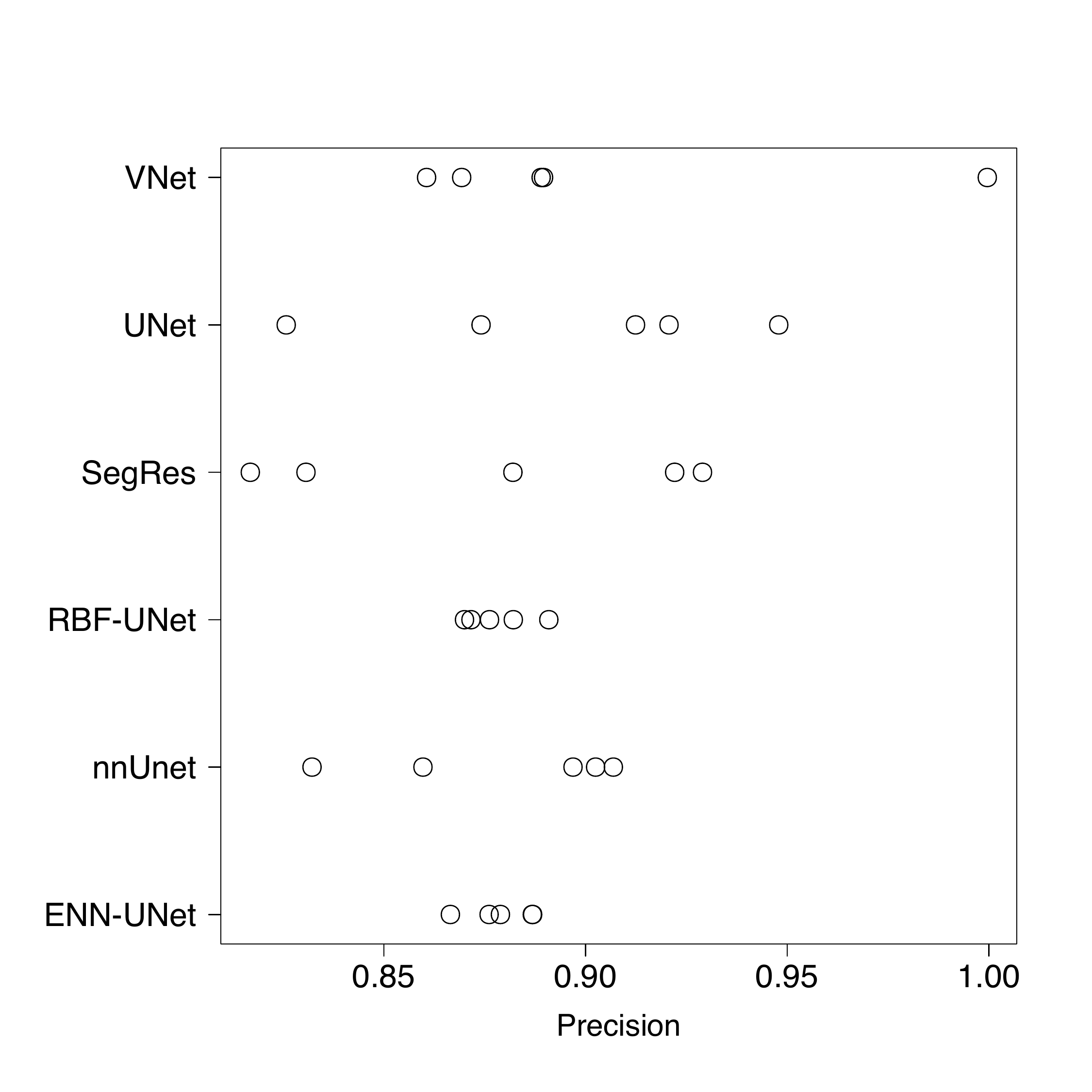}}
\caption{Values of the Dice score (a), sensitivity (b) and precision (c) for five runs of the six methods. \label{fig:dotplots}}
\end{figure}

 The means and standard deviations of the Dice score, sensitivity and precision over five runs with random initialization for the six methods are shown in Table~\ref{tab:compar}, and the raw values are plotted in Figure \ref{fig:dotplots}. We can see that ENN-UNet and RBF-UNet achieve, respectively, the highest and the second highest mean Dice score. A Kruskal-Wallis test performed on the whole data concludes to a significant difference between the distributions of the Dice score for the six methods (p-value = 0.0001743), while the differences are not significant for sensitivity (p-value = 0.2644) and precision (p-value = 0.9496). Table \ref{tab:conover} shows the results of the Conover-Iman test of multiple comparisons \cite{conover79}\cite{dinno17} with Benjamini-Yekutieli adjustment \cite{BY01}. We can see that the differences between the Dice scores obtained by ENN-UNet and RBF-UNet on the one hand, and the four other methods on the other hand are highly significant (p-values $< 10^{-2}$),  while the difference between ENN-UNet and RBF-UNet is only weakly significant (p-value = 0.0857).

\begin{table}
\caption{Means and standard deviations (over five runs) of the performance measures for ENN-UNet, RBF-UNet and four reference methods. The best result is shown in bold, and the second best is underlined.}
\centering
\label{tab:compar}
\begin{tabular}{ccccccc}
\hline
 Model  &\multicolumn{2}{c}{Dice score} &\multicolumn{2}{c}{Sensitivity}&  \multicolumn{2}{c}{Precision}\\
 \cline{2-7} 
 &Mean&SD&Mean&SD&Mean&SD\\
\hline
UNet \cite{kerfoot2018left} & 0.753&0.054& 0.782 &0.048&\underline{0.896}&0.047\\
nnUNet \cite{isensee2018nnu} & 0.817 &0.008& \textbf{0.838}&0.028  &0.879&0.032 \\
VNet \cite{milletari2016v}&0.820&0.016&0.831&0.021 & \textbf{0.901}&0.056 \\
SegResNet \cite{myronenko20183d}& 0.825&0.015& \underline{0.832}&0.042&0.876&0.051 \\
\hline
ENN-UNet&\textbf{0.846}&0.002&0.830&0.004&0.879&0.008\\
RBF-UNet&\underline{0.839}&0.003&0.824&0.001&0.879&0.008\\

\hline
\end{tabular}
\end{table}

\begin{table}
    \centering
       \caption{Conover-Iman test of multiple comparisons between the Dice scores obtained by the six models: t-test statistics and p-values. P-values less than 0.01 are printed in bold. \label{tab:conover}}
    
    \begin{tabular}{lccccc}
    \hline
        &    ENN-UNet  &   nnUnet &  RBF-UNet  &   SegResNet  &     UNet\\
        \hline
  nnUnet &   6.759 & & & & \\
         &    \textbf{0.0000} & & & & \\
RBF-UNet &   2.156 & -4.602 & & &  \\
         &     0.0857 &   \textbf{0.0004}& & &  \\
  SegResNet &   5.349 & -1.410 &  3.193 & &   \\
         &    \textbf{0.0001}  &   0.3282 &   \textbf{0.0088}& &   \\
    UNet &   10.283 &  3.524  & 8.127 &  4.934 &    \\
         &    \textbf{0.0000}  &  \textbf{0.0043} &   \textbf{0.0000}  &  \textbf{0.0002}&    \\
    VNet &   6.054&  -0.705 &  3.898  & 0.705 & -4.229\\
         &    \textbf{0.0000}  &   0.8091  &  \textbf{0.0019}   &  0.8669  &  \textbf{0.0009}\\
         \hline
    \end{tabular}
\end{table}

\begin{figure}
\centering
\includegraphics[scale=0.45]{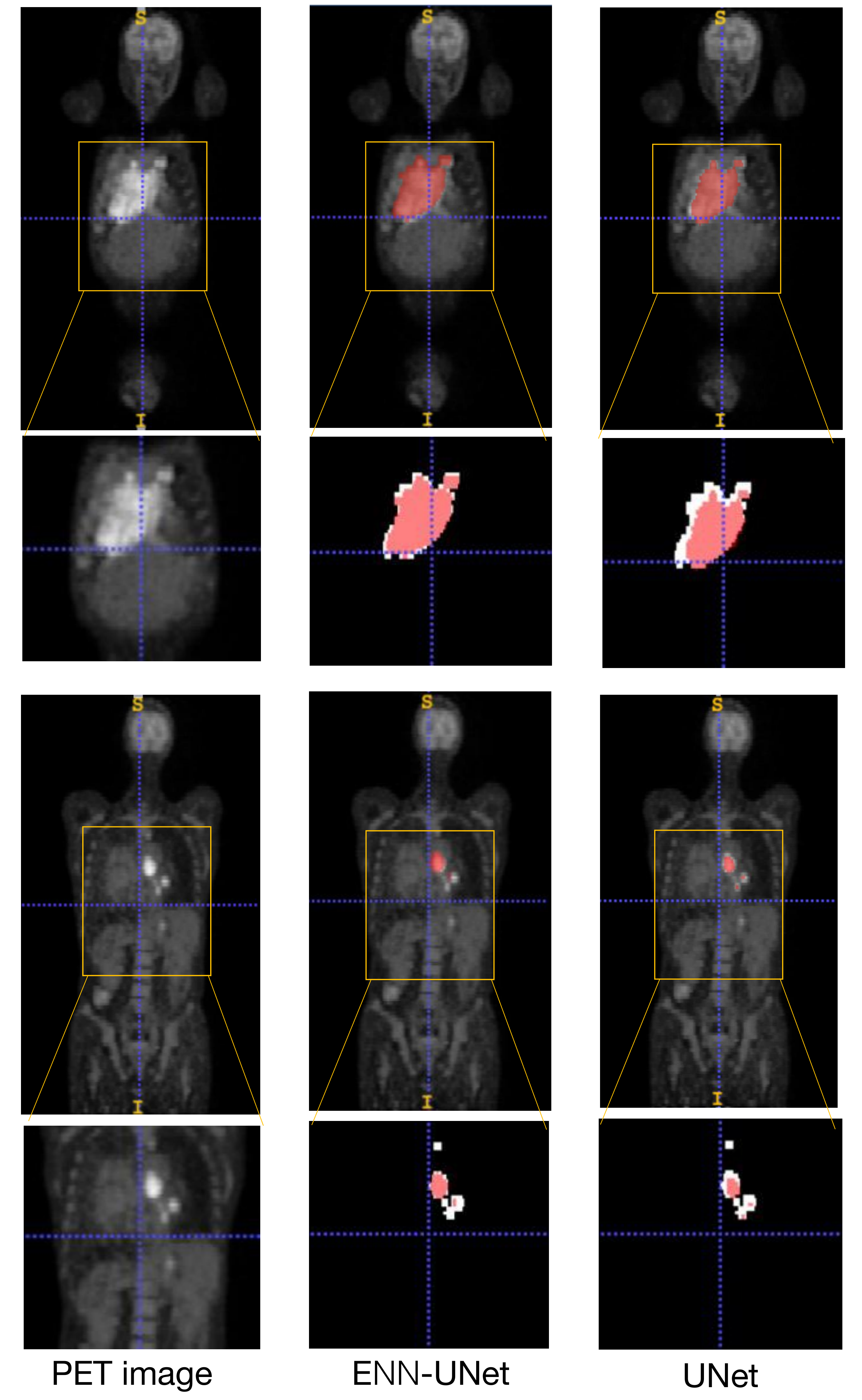}
\caption{Two examples of segmentation results by ENN-UNet and UNet. The first and the second row are, respectively, 
 representative of large and isolated small lymphomas. The three columns correspond, from left to right, to the PET images and the  segmentation results obtained by ENN-UNet and UNet. The white and red region represent, respectively, the ground truth and the segmentation result. \label{fig:seg_result}}
\end{figure}

Figure \ref{fig:seg_result} shows two examples of segmentation results obtained by ENN-UNet and UNet, corresponding to large and isolated lymphomas. We can see, in these two examples, that UNet is more conservative (it correctly detects only a subset of the tumor voxels), which may explain why it has a relatively high precision. However, the tumor regions predicted by ENN-UNet better overlap the ground-truth tumor region, which is also reflected by the higher Dice score. 

\new{
\subsection{Comparative analysis: calibration} 
\label{subsec:calibration}

Besides segmentation accuracy, another important issue concerns the quality of uncertainty quantification. Monte-Carlo  dropout (MCD) \cite{gal2016dropout} is a a state-of-the-art technique for improving uncertainty quantification capabilities of deep networks. In this section, we compare the  ECE \eqref{eq:ECE} achieved by  UNet (the baseline),  SegResNet (the best alternative method found in Section \ref{subsec:state-of-art}), and our proposals: ENN-UNet, and RBF-UNet, with and without MCD. For the four methods, the dropout rate was set to  0.5 and the sample number was set to 20; we averaged the 20 output probabilities (the pignistic probabilities for the two evidential models) at each voxel as the final output of the model. 

The results are reported in Table \ref{tab:ECE}. We can see that MCD enhances the segmentation performance (measured by the Dice index) of UNet et SegResNet, and improves the calibration of all methods, except SegResNet. Overall, the smallest average ECE is achieved by  RBF-UNet and ENN-UNet with MCD, but the standard deviations are quite large. A Kruskal-Wallis test concludes to a significant difference between the distributions of ECE for the eight methods (p-value = 0.01). The p-values of the Conover-Iman test of multiple comparisons with Benjamini-Yekutieli adjustment reported in Table \ref{tab:conover-calib} show significant differences between the ECE of RBF-UNet with MCD one the one hand, and those of RBF-UNet without MCD, SegResNet with MCD, and  UNet without MCD on the other hand. We also tested the pairwise differences between the ECE values obtained by RBF-UNet and ENN-UNet with MCD on the one hand, and UNet with and without MCD as well as SegResNet with and without MCD on the other hand using the Wilcoxon rank sum test. The corresponding p-values are shown in Table \ref{tab:wilcox}. We find significant differences between the ECE  RBF-UNet with MCD and those of the other methods, but only a weakly significant difference between ENN-UNet with MCD and UNet without MCD. In summary, there is some  evidence that MCD improves calibration, even for evidential models, and that the  best calibration is achieved by the  RBF-UNet model, but this evidence is not fully conclusive due to the limited size of the dataset; our findings  will  have to be confirmed by further experiments with larger datasets.

\begin{table}
\centering
\caption{\new{Means and standard deviations (over five runs) of the Dice score and ECE for UNet, SegResNet, ENN-UNet andRBF-UNet, with and without MCD. The best results are shown in bold, the second best are underlined.}}
\new{
    \begin{tabular}{cccccc}
    \hline
     Model  &\multicolumn{2}{c}{Dice score} &\multicolumn{2}{c}{ECE(\%)}\\
 \cline{2-3} 
 \cline{4-5} 
 &Mean&SD&Mean&SD\\
  \cline{1-5} 
    UNet       &0.754&0.054&  2.22&0.205\\
    SegResNet &0.825&0.015&  1.97&0.488\\
    ENN-UNet   & \textbf{0.846} & 0.002 &  1.99 &0.110\\
    RBF-UNet   &0.839&0.003&  2.12&0.028\\
    \hline
    UNet with MC &0.828&0.005&1.93&0.337\\
    SegResNet with MC &\underline{0.844}&0.009&2.53&0.973\\
    ENN-UNet with MC& 0.841&0.003&\underline{1.53}&0.075 \\
    RBF-UNet with MC &0.840&0.003&\textbf{1.52}&0.041\\
    \hline
    \end{tabular}
    }
    \label{tab:ECE}
\end{table}

\begin{table}
    \centering
       \caption{\new{Conover-Iman test of multiple comparisons between the ECE  obtained by UNet, SegResNet, ENN and RBF, with and without MCD: t-test statistics and p-values. P-values less than 0.01 are printed in bold.} \label{tab:conover-calib}}
    \new{
    \begin{tabular}{lccccccc}
    \hline
&       ENN  &   ENN-MC   &     RBF  &   RBF-MC &  SegRes  & SegRes-MC& UNet\\
\hline
  ENN-MC &   0.926 &&&&&&\\
         &     1.0000 &&&&&&\\
     RBF &  -1.191 & -2.118 &&&&&\\
         &     0.7403 &    0.2892 &&&&&\\
  RBF-MC &   2.812 &  1.886 &  4.004 &&&&\\
         &     0.1145 &    0.3419 &   \textbf{0.0095} &&&&\\
SegRes &   0.695 & -0.232 &  1.886 & -2.117 &&&\\
         &     1.0000 &    1.0000 &    0.3761 &    0.3305 &&&\\
SegRes-MC &  -0.860 & -1.787 &  0.331 & -3.673  &-1.555 &&\\
         &     1.0000 &    0.3530 &    1.0000 &   \textbf{0.0159} &    0.4756 &&\\
    UNet &  -1.357 & -2.283 & -0.165 & -4.169 & -2.051 & -0.496 & \\
         &     0.6337 &    0.2677 &    1.0000 &   \textbf{0.0119} &    0.2962 &    1.0000 & \\
 UNet-MC &   0.430 & -0.496 &  1.621 & -2.382 & -0.265  & 1.290  &  1.787\\
         &     1.0000 &    1.0000  &   0.4507  &   0.2564  &   1.0000  &   0.6667  &  0.3824\\
 \hline
    \end{tabular}
    }
\end{table}   

\begin{table}
    \centering
       \caption{\new{P-values for the Wilcoxon rank sum test applied to the comparison of ECE obtained by ENN-UNet and RBF UNet with MCD on the one hand, and the four other methods on the other hand (UNet and SegResNet with and without MCD).} \label{tab:wilcox}}
    \new{
    \begin{tabular}{lcccc}
    \hline
 & UNet &   UNet-MC &  SegRes  & SegRes-MC\\
    \hline
ENN-MC& 0.095 &0.67 &0.69& 0.31\\
RBF-MC& 0.0079 &0.012 &0.055 &0.0079\\
\hline
    \end{tabular}
    }
\end{table} 

}

\section{Conclusion}
\label{sec:conc}

An evidential framework for segmenting lymphomas from 3D PET-CT images with uncertainty quantification has been proposed in this paper. Our architecture is based on the concatenation of a UNet, which extracts high-level features from the input images, and an evidential segmentation module, which computes output mass functions for each voxel. Two versions of this evidential module, both involving prototypes, have been studied: one is based on the ENN model initially proposed as a stand-alone classifier in \cite{denoeux2000neural}, while the other one relies on an RBF layer and the addition of weight of evidence. The whole model is trained end-to-end by minimizing the Dice loss. The initialization of prototypes has been shown to be a crucial step in this approach. The best method found has been to pre-train a UNet with a softmax output layer, initialize the prototype with the $k$-means algorithm in the space of extracted features, train the evidential layer separately, and fine-tune the whole network. Our model has been shown to outperform the baseline UNet model as well as other state-of-the-art segmentation method on a dataset of 173 patients with lymphomas. \new{Preliminary results also suggest the outputs of the evidential models (in particular, the one with an RBF layer) are better calibrated and that calibration error can be further decreased by Monte Carlo dropout. These results, however, will have to be confirmed by further experiments with larger datasets. }

This work can be extended in many directions. One of them is to further evaluate the approach by applying it to other medical image segmentation problems. One of the potential problems that may arise is related to the dimensionality of the feature space. In the application considered in this paper, good results where obtained with only two extracted features. If some other learning tasks require a much larger number of features, we may need a much higher number of prototypes and learning may be slow. This issue could be addressed by adapting the loss function as proposed, e.g., in \cite{Hryniowski20}. We also plan to further study the calibration properties of the belief functions computed by our approach \new{(using calibration measures specially designed for belief functions)}, as well as the novelty detection capability of our model.

\section*{Acknowledgements} 

This work was supported by the China Scholarship Council (No. 201808331005). It was carried out in the framework of the Labex MS2T, which was funded by the French Government, through the program ``Investments for the future'' managed by the National Agency for Research (Reference ANR-11-IDEX-0004-02)

\section*{References}

\end{document}